%% file: main.tex
\pdfoutput=1
\documentclass[letterpaper]{article}
\usepackage[preprint]{aaai2027}
\usepackage[hyphens]{url}
\usepackage{graphicx}
\usepackage{natbib}
\usepackage{caption}
\usepackage{amsmath}
\usepackage{amssymb}
\usepackage{amsthm}
\usepackage{booktabs}
\usepackage{multirow}
\usepackage[ruled,vlined]{algorithm2e}

\renewcommand{\topfraction}{0.88}
\renewcommand{\bottomfraction}{0.65}
\renewcommand{\textfraction}{0.10}
\renewcommand{\floatpagefraction}{0.80}
\newtheorem{proposition}{Proposition}

\newtheorem{lemma}{Lemma}
\newtheorem{remark}{Remark}

\title{Reframing Video Inverse Tone Mapping as Blind Posterior\\Inference via Latent Luminance Supervision}
\makeatletter
\def\@fnsymbol#1{\ensuremath{\ifcase#1\or \dagger\or \ddagger\or \ast\or \mathsection\or \mathparagraph\else\@ctrerr\fi}}
\makeatother
\author{
    Hanwool Sul\textsuperscript{\rm 1,2},
    Chang-Hoon Jeong\textsuperscript{\rm 1},
    Kisung Shin\textsuperscript{\rm 1},
    Bomseok Park\textsuperscript{\rm 2},\\
    Donghoon Cha\textsuperscript{\rm 2},
    Hansl von Kwon\textsuperscript{\rm 2},
    Ku DoHyung\textsuperscript{\rm 2},
    Byoung-Tak Zhang\textsuperscript{\rm 1}\corresponding
}
\affiliations{
    \textsuperscript{\rm 1}Seoul National University\\
    \textsuperscript{\rm 2}STUDIO FREEWILLUSION
}

\begin{document}
\maketitle

\begin{abstract}
High-dynamic-range (HDR) displays are now common, but most video content, such as archival footage, broadcast material, and the output of generative video models, exists only in standard dynamic range (SDR). The need is sharpest for generated video, which has no HDR master at all and is graded alongside live-action and visual-effects material finished in HDR. Converting SDR to HDR means undoing a tone-mapping operation that was never recorded, and clipping destroys the radiance wherever it acts, so HDR recovery is a blind inverse problem. Methods built on generative video priors fill clipped regions with plausible structure, but their reconstructions are systematically biased: in every public method we test, saturated content leans too bright or too dark, and the direction of the lean moves with the tone operator behind the input. We call this failure the \emph{luminance error}, measure both its magnitude and its sign, and make it the focus of this work. Reframing ITM as posterior inference under an unknown operator locates the failure in two places and yields two corrections. A fixed affine functional of the video-diffusion latent reads log-luminance without decoding; on that coordinate an asymmetric loss selects an expectile of the posterior, applied only where clipping leaves luminance unconstrained. We also synthesize training SDR through an explicit family of tone operators, so the model marginalizes over what the observation cannot show. Trained this way, our luminance error stays flatter across operators and smaller on average than any public method, and the full system, which we call ReITM, achieves the best perceptual quality on all three benchmarks.

\end{abstract}

\begin{figure}[t!]
\centering
\includegraphics[width=0.97\linewidth]{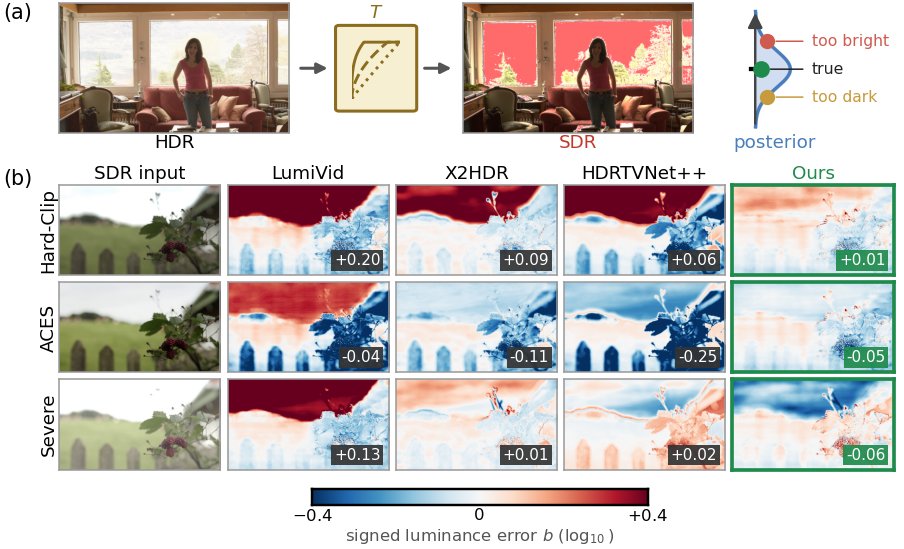}
\caption{\textbf{(a)} Red marks clipped pixels; there the posterior is wide (right). \textbf{(b)} One clip under three operators: signed luminance error after median-luminance alignment, bias $b$ inset. Only the operator changes between rows.}
\label{fig:concept}
\end{figure}

\begin{figure*}[t]
\centering
\includegraphics[width=\textwidth]{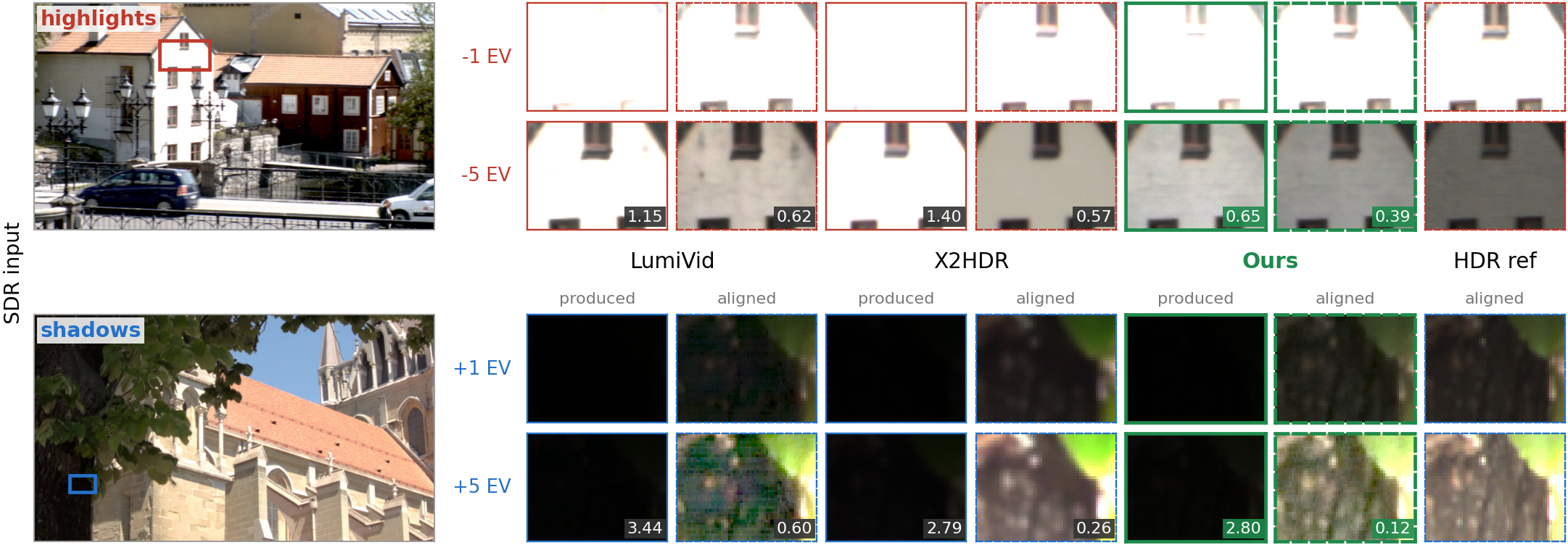}
\caption{The same crop re-exposed. Each tile is linear $2^{\text{EV}}$ with $2.2$ gamma, so highlights that kept detail separate from highlights filled with a constant. Solid rows are the produced scale, dashed rows the median-luminance alignment of \S5. Insets: region RMSE. Black tiles: no method matches the reference's absolute scale.}
\label{fig:teaser}
\end{figure*}

\section{Introduction}
High-dynamic-range (HDR) displays are now commonplace, but most video exists only as an eight-bit standard-dynamic-range (SDR) signal. The need is sharpest for generated video: archival footage was at least mastered in SDR as a finished product, whereas today's state-of-the-art video generators emit eight-bit SDR frames natively \citep{hunyuanvideo,wan,ltx2}, with no HDR master behind them, and their output is increasingly graded alongside live-action and visual-effects material finished in HDR. Where an HDR delivery path exists for such a model, it is an inverse-tone-mapping adapter applied to the SDR output after generation \citep{lumivid}, the very problem studied here. Recovering HDR from SDR, or \emph{inverse tone mapping} (ITM), is therefore not only restoration for legacy content but a prerequisite for using generated video in delivery. We model an SDR signal as scene radiance passed through a \emph{tone-mapping operator} (TMO) $T$ that compresses the dynamic range and clips whatever exceeds the white point, $\mathrm{SDR}=\mathrm{clip}(T(\mathrm{HDR}))$. Two properties define the task. $T$ is never recorded, and for generated video no physical operator existed at all, so a model tuned to one operator has no stable structure to exploit: the setting demands behavior independent of which operator produced the input. And wherever the signal clips, the observation is locally constant, so the lost radiance must be generated.

A pretrained video-diffusion model is the natural engine for that generation, and recent ITM systems build on such priors, treating the remaining fidelity gap as the price of generation \citep{lumivid}. Some also train across multiple tone operators \citep{lumaflux,realrep}, but to our knowledge no generative video ITM method has been tested on tone operators excluded from its training family, so operator robustness is asserted rather than measured (\citet{realrep} tests unseen degradations, but for single-image regression). We measure that gap. After a global exposure match to the reference, we read the log-luminance error through its root mean square and its signed mean, which says which way the reconstruction leans. That mean is not zero and not erratic: it moves with the operator behind the input, in a direction characteristic of each method, from $+0.09$ for one public model under hard clipping to $-0.26$ for another under an ACES rendering (Fig.~\ref{fig:tilt}, Table~\ref{tab:tilt}). A well-calibrated hallucination has no reason to lean; a lean that tracks the operator is a bias. Two facts make it a problem in its own right. It is invisible to standard practice: quality metrics discard the error's sign, and aligning a prediction to the reference absorbs the quantity ITM must get right (\S2). It also cannot be repaired afterwards, since it is a property of the conditional mean the model learned (\S3).

We make both observations precise by reframing ITM as posterior inference under an unknown operator, a blind inverse problem in the sense of blind deconvolution \citep{levin2009}:
\begin{equation}
p(\mathrm{HDR}\mid\mathrm{SDR}) \;\propto\; p(\mathrm{HDR})\;\mathbb{E}_{T\sim p(T)}\!\big[\,p(\mathrm{SDR}\mid \mathrm{HDR},T)\,\big].
\label{eq:posterior}
\end{equation}
The prior is the pretrained diffusion model, and the unknown operator enters the posterior twice. It enters the likelihood: a model trained under one operator learns the wrong conditional for every other, which is where the operator dependence comes from. It also enters the estimate, since the model commits to a single value drawn from the posterior, and wherever clipping leaves that posterior wide, the value it commits to decides whether the output leans bright or dark. We name this failure the \emph{luminance error}. Our main contributions, which address this failure through both its measurement and our two-part correction, are as follows:

\begin{enumerate}
\item \textbf{The luminance error.} A signed, operator-resolved metric and evaluation protocol. It exposes a systematic bias where reconstructions lean too bright or too dark depending on the input operator, a failure invisible to standard magnitude-only, curve-aligned evaluations.
\item \textbf{A luminance coordinate and a derived correction (\S4.1).} A fixed affine functional of the frozen latent reads log-luminance without decoding ($R^{2}=0.96$ on clips never used to fit it, against $0.27$ for the best single principal component) and advances one decade per decade of exposure. On this coordinate we train with an observability-gated asymmetric loss whose optimal action is a posterior expectile, reducing the bias on all six evaluated operators.
\item \textbf{Operator-marginalized training (\S4.2).} Synthesizing SDR through an explicit five-operator family marginalizes the likelihood; at a corpus scale comparable to the strongest baseline the full system, ReITM, yields the lowest luminance error across operators (mean RMSE $0.141$ versus $0.179$--$0.341$) and first place on all three benchmarks (CVVDP $7.690/8.567/6.879$).
\end{enumerate}

\begin{figure*}[t]
 \centering
 \includegraphics[width=\textwidth]{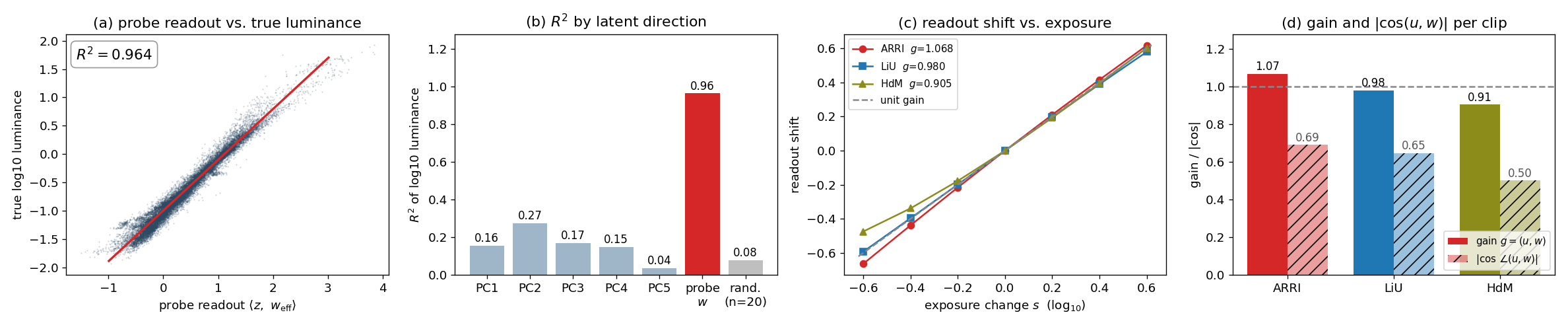}
 \caption{\textbf{(a)} Probe readout against true log-luminance, one point per latent token: $R^{2}=0.96$ on clips the probe was not fitted on. \textbf{(b)} The same $R^{2}$ for other latent directions. No principal component exceeds $0.27$, so luminance is carried by one specific low-variance direction, not by the latent's dominant structure. \textbf{(c)} Readout shift under a $\pm2$\,EV sweep: one decade of readout per decade of exposure to within $10\%$, with no per-clip fit. \textbf{(d)} Per clip, the gain $g$ against $|\cos\angle(u,w)|$. Here $g$ stays near one while $|\cos\angle(u,w)|$ stays well below it, so unit gain does not require alignment with the exposure direction.}
\label{fig:probe}
\end{figure*}

\section{Related Work}
\paragraph{Generative inverse tone mapping.}
ITM began with hand-designed expansion operators \citep{banterle2006}, moved to CNN regression \citep{eilertsen2017}, and was extended to video \citep{kim2019}. Because clipped regions carry no signal, recent work replaces regression with generative priors: \citet{lumivid} aligns a logarithmic encoding with a pretrained video-diffusion latent but leaves the training operator unspecified; \citet{mevm} merges synthesized exposure brackets; \citet{diffhdr} re-exposes LDR video with a video-diffusion prior; \citet{lumaflux} augments with multiple operators yet evaluates on the same ones. Regression-based video ITM has also exploited temporal information \citep{ye2024}. A separate line applies diffusion priors to inverse problems with a known operator \citep{dps,pigdm} or estimates it jointly with the signal \citep{blinddps,gibbsddrm}. Both lines fix the operator, estimate it entirely, or leave it implicit, whereas we split it by identifiability (clipped regions identify nothing, \S3) and marginalize the unidentifiable part at training time.

\paragraph{Treatment of the tone operator.}
\citet{hdrtvnet} concedes that SDR can be produced in many ways, yet trains on a single conversion style, and its extension \citep{hdrtvnetpp}, which we run as a baseline, still fixes the formation model at training time. Others sample random camera curves as an augmentation, without relating the choice to an identifiability argument \citep{lediff}, work from RAW rather than from a tone-mapped signal \citep{x2hdr}, or keep the operator as an argument inferred at test time from a learned degradation embedding \citep{realrep}, the closest problem statement to ours, though posed for single-image regression. We keep the formation view but treat the operator as a random variable and marginalize it during training, so the argument is removed rather than inferred.

\paragraph{Evaluating HDR reconstruction.}
Standard HDR metrics such as PU21-encoded PSNR/SSIM \citep{pu21} and visual-difference predictors \citep{cvvdp} report error magnitude and discard its sign, and common protocols align the prediction to the reference before scoring (\citet{mevm} uses both an affine match and a re-fitted response curve). A single global factor only fixes overall exposure, which no method can identify from an SDR observation (\S3); matching it is legitimate. Re-fitting an affine map or a response curve instead estimates the tone transform ITM is asked to recover, so a reconstruction that leans the wrong way still scores well. Directional error is therefore invisible to curve-aligned, magnitude-only evaluation. We never re-fit a response curve, report exposure-only alignment as primary, and introduce the luminance error to isolate the component existing metrics discard.

\section{Analysis: What the Observation Reveals}
We reframe ITM as the blind inverse problem of Eq.~\eqref{eq:posterior}, with $H$ the scene HDR, $S$ the observed SDR, and $T\!\sim\!p(T)$ the unknown operator. Three properties drive the method; proofs are in the supplement.

\begin{figure*}[t]
 \centering
 \includegraphics[width=\textwidth]{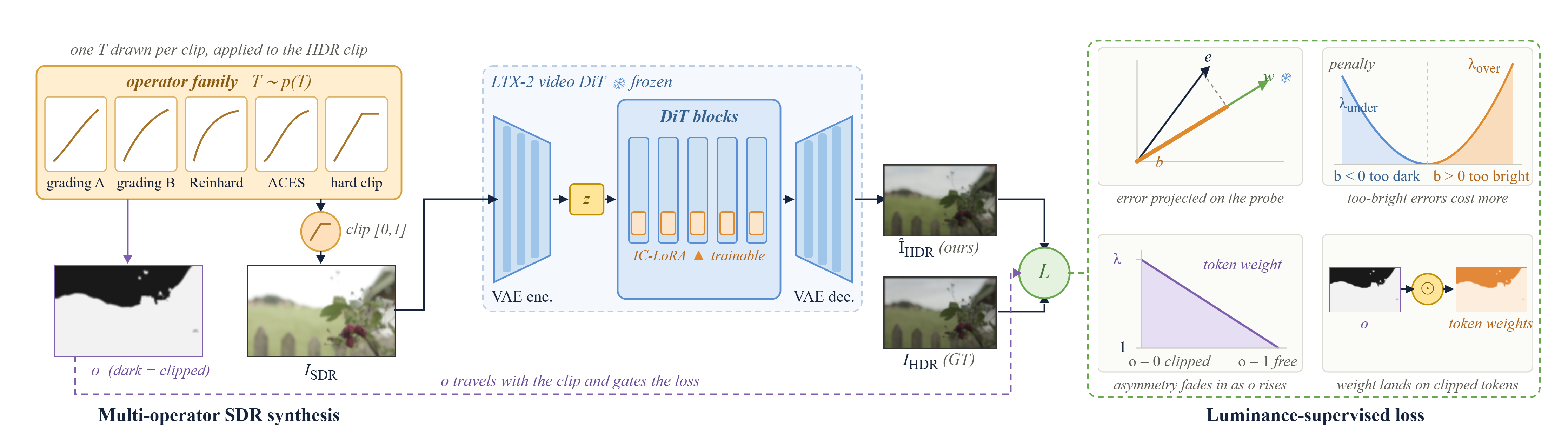}
 \caption{Method overview on a real evaluation clip. \textbf{Left:} One operator per clip, drawn from the family of \S4.2, is applied to the HDR source and clipped, yielding $I_{\mathrm{SDR}}$ and the observability map $o$ (dark $=$ clipped). \textbf{Center:} The SDR conditions a VAE latent $z$; only the in-context LoRA adapters are trained. \textbf{Right:} The loss (\S4.1).}
 \label{fig:arch}
\end{figure*}

The observation identifies the operator only where it does not clip. Per pixel it is $S_i=\min(f(\alpha H_i),\,s_{\max})$ for a monotone tone curve $f$ and exposure $\alpha$.
\begin{proposition}[Partial identifiability]
On unsaturated pixels ($S_i<s_{\max}$) the observation determines $H_i$ once the operator is fixed; on saturated pixels the likelihood is flat above the clip threshold $c(T)$ \citep{zhang2004saturated}, so neither the operator nor the radiance is identifiable. Identifiability degrades continuously as the clipped fraction grows.
\label{prop:ident}
\end{proposition}
We summarize this with a per-token \emph{observability} $o\in[0,1]$, near $1$ where the measurement pins the value and near $0$ where the posterior reduces to the prior. The design rule follows: read the observation where $o$ is high, generate under operator uncertainty where it is low.

The operator dependence is a variance term, not noise. Let $\varepsilon=\hat H-H$; conditioning on the unknown operator and applying the law of total variance,
\begin{equation}
\mathrm{Var}(\varepsilon)=\underbrace{\mathbb{E}_T[\mathrm{Var}(\varepsilon\mid T)]}_{\text{within-operator}}+\underbrace{\mathrm{Var}_T(\mathbb{E}[\varepsilon\mid T])}_{\text{between-operator}}.
\label{eq:lotv}
\end{equation}
The two terms respond differently to training. A model trained on one operator $T_0$ drives its within-operator conditional mean toward zero yet leaves $\mathbb{E}[\varepsilon\mid T]\ne 0$ for every $T\ne T_0$, so the between-operator term persists and grows with the mismatch (verified in \S5). Because it is a property of the trained conditional mean, no test-time alignment removes it: only marginalizing the operator during training drives $\mathbb{E}[\varepsilon\mid T]$ to zero on average across the family, the same mechanism domain randomization uses against an unknown simulator gap \citep{tobin2017}. This holds only over $\mathrm{supp}\,p(T)$, a range condition the boundary analysis of \S5 tests on our own model.

We quantify the failure with the \emph{luminance error}. After a median-luminance alignment that removes the one degree of freedom ITM cannot fix, let $e=\log_{10}\hat Y-\log_{10}Y$ per pixel. Its root mean square $L=\sqrt{\smash[b]{\mathbb{E}[e^{2}]}}$ says how wrong the luminance is; its mean, the signed bias $b=\mathbb{E}[e]$, says which way it leans. They cannot cancel: $L^{2}=b^{2}+\mathrm{Var}(e)$, so a nonzero $b$ is evidence of systematic bias.

Both branches of the method must read luminance from the latent inside the training loop, without decoding. We fit one affine readout of the latent, once, by ridge least squares on (latent token, $\log_{10}Y$) pairs, and freeze it with the backbone: with $\mathrm{Enc}$ the frozen video autoencoder, $z$ any latent token and $s=\log_{10}\alpha$ the log-exposure,
\begin{equation}
r(z):=\langle z,w\rangle+\beta,\qquad u:=\left.\frac{\partial \mathrm{Enc}}{\partial s}\right|_{s=0},\qquad g:=\langle u,w\rangle.
\label{eq:probe}
\end{equation}
Exposure acts on the log-encoded input as one additive offset at every pixel away from the encoding's toe and floor, so it displaces the latent along $u$. The readout $w\in\mathbb{R}^{128}$ is a linear functional read off that displacement, and it is normalized \emph{on} $u$, not aligned \emph{with} it.
\begin{proposition}[Exposure-affine luminance readout]
Let $\mathrm{Enc}$ be differentiable in $s$ on the working range. Then for any fixed $(w,\beta)$,
$r(\mathrm{Enc}(10^{s}H))=r(\mathrm{Enc}(H))+g\,s+O(s^{2})$ with the single scalar $g=\langle u,w\rangle$, so only the component of the displacement along $w$ reaches the readout. Unit gain $g=1$ then constrains only the product $\lVert u\rVert\lVert w\rVert\cos\angle(u,w)$ and does \emph{not} require $w\parallel u$ (supplement). One consequence is used below: the intercept $\beta$ cancels in differences, $r(\hat z)-r(z)=\langle\hat z-z,w\rangle$, so for $g>0$ the sign of that projection is the sign of the readout's luminance error, the only quantity the loss of \S4.1 consumes.
\label{prop:probe}
\end{proposition}
What is empirical is whether the fitted $(w,\beta)$ achieves a high $R^{2}$ and unit gain (Fig.~\ref{fig:probe}). As a regressor, the probe explains $R^{2}=0.96$ of per-token log-luminance variance on $16$ clips never used to fit it ($0.95$ with the stored intercept $\beta$ and no refit). As an exposure readout, a global $\pm2$\,EV sweep ($\pm0.6$ decades) on three clips gives $g=0.905$--$1.068$, with the readout affine in $s$ ($R^{2}\ge0.995$): one decade of readout per decade of exposure, to within $10\%$, with no per-clip fit. Luminance is carried by a specific low-variance direction: no leading principal component or random direction comes close (Fig.~\ref{fig:probe}b). The fit transfers only partially to a second, independently trained autoencoder, so we scope the readout as demonstrated on our backbone (supplement).
The gain is a pairing: $g=\lVert u\rVert\lVert w\rVert\cos\angle(u,w)$, and the measured $u$ sits $46^{\circ}$--$59^{\circ}$ from $w$ while $g$ stays near one (Fig.~\ref{fig:probe}d). Alignment is the wrong target: projecting on the exposure direction is the worse of the two readouts, and $u$ is only approximately clip-independent (supplement).

\section{Method}

Proposition~\ref{prop:ident} splits the method in two (Fig.~\ref{fig:arch}): where identification succeeds, the observation supplies supervision; where it fails, the posterior is wide and the operator unrecoverable, so we correct the \emph{direction} of the estimate (\S4.1) and train against the whole family (\S4.2). Both branches preserve the pretrained prior. The two branches answer different halves of the diagnosis. The family of \S4.2 attacks the between-operator term of Eq.~\eqref{eq:lotv}; the directional correction of \S4.1 attacks the bias $b$ that survives within an operator. \S5 ablates them separately.

\subsection{The luminance coordinate and the asymmetric action}
The model returns one reconstruction, a single value it commits to from the posterior, an \emph{action}. Where clipping leaves the posterior wide (low $o$), that choice fixes the sign of the error. We move the action along the luminance coordinate of Proposition~\ref{prop:probe} by reweighting the loss by the sign the probe reads: the velocity error $\delta=\hat v-v$ is contracted with the probe to give a per-token signal
\begin{equation}
\text{lum\_dir}=-\langle \delta,\,w_{\text{eff}}\rangle,
\end{equation}
where $w_{\text{eff}}=w/\sigma$ is the fitted readout of Proposition~\ref{prop:probe} in the trainer's raw latent basis ($\sigma$ the standardization scale of the fit; no intercept enters, since the probe is contracted with a residual), and the sign is chosen so $\text{lum\_dir}>0$ means the prediction is too bright. The contraction is taken on the velocity residual but reads the reconstruction. Since $x_t=x_0+t\,v$, the predicted clean latent obeys $\hat x_0-x_0=-t\,\delta$, so $\text{lum\_dir}=\langle\hat x_0-x_0,\,w_{\text{eff}}\rangle/t$. The factor $t$ is positive and changes only the scale, so at every noise level the sign is the sign of the readout's luminance error on the reconstruction. Only the sign of this scalar enters the loss, so the readout's scale, its intercept and its measured $10\%$ gain spread leave the correction unchanged; the residual is only read along $w_{\text{eff}}$, never displaced along it. The gate needs a number for $o$. We estimate the response the operator actually applied without knowing its parameters: within a clip we bin log radiance into $48$ bins, take the median SDR level in each, and read the slope $\partial S/\partial\log_{10}Y$ of that empirical curve. The slope is how far the observation moves per stop, so it vanishes where the signal clips. We rescale it to $[0,1]$ across the clip and average it inside each latent token. The estimate presumes no functional form for $T$, and it is read off the training pair $(H,S)$, so $o$ exists only during training: it weights the loss and never enters the network, which at inference receives the SDR alone. The loss weights each token asymmetrically and gates the asymmetry by observability:
\begin{equation}
\lambda(\text{lum\_dir})=\begin{cases}\lambda_{\text{over}}=2.0 & \text{lum\_dir}>0\\ \lambda_{\text{under}}=1.0 & \text{lum\_dir}\le 0\end{cases}
\end{equation}
\begin{equation}
\text{tok\_w}=1+\big(\lambda(\text{lum\_dir})-1\big)\,(1-o).
\label{eq:gate}
\end{equation}

Two properties, proved in the supplement, make this principled rather than heuristic. First, the Bayes-optimal action of the asymmetric quadratic loss is the $\tau$-expectile of the projected posterior $\langle x_0,w_{\text{eff}}\rangle$, with $\tau=\lambda_{\text{under}}/(\lambda_{\text{over}}+\lambda_{\text{under}})=1/3$. The first-order condition of the conditional risk, $\lambda_{\text{over}}\,\mathbb{E}[(a-X)^{+}]=\lambda_{\text{under}}\,\mathbb{E}[(X-a)^{+}]$, is the defining equation of that expectile, and monotonicity of its two sides gives uniqueness. This is an instance of the correspondence between a scoring rule and the functional it elicits \citep{gneiting2011}, and the same construction learns expectiles in reinforcement learning \citep{iql2022}. The loss does not change the posterior, only the point read from it. Second, the gate is optimal in the two limits set by Proposition~\ref{prop:ident}: on observed tokens ($o\!\to\!1$) the posterior is pinned and asymmetry only adds risk, while on clipped tokens ($o\!\to\!0$) an asymmetry in the direction of the misspecification shift cancels it. The asymmetric-$L_2$ form keeps the gradient structure of the diffusion objective, so the correction is a per-token reweighting (Algorithm~\ref{alg:train}). The observability gate determines \emph{where} the asymmetry acts; the sign of the misspecification shift is resolved analytically only when the test operator is known to dominate the family in one direction (supplement), and which direction an unseen operator takes is not known in advance, so the direction of the asymmetry and the ratio $2.0/1.0$ are selected on validation. Against multi-operator training alone, adding this term reduces $|b|$ on every evaluation operator (\S5).

\begin{algorithm}[t]
\SetAlgoLined
\DontPrintSemicolon
\SetCommentSty{algcomment}
\footnotesize
\SetKwInOut{Input}{input}\SetKwInOut{Param}{fixed}
\Input{HDR clip $H$; LoRA parameters $\theta$; step size $\eta$}
\Param{family $\mathcal{T}=\{T_1,\dots,T_5\}$; readout $w\in\mathbb{R}^{128}$; frozen VAE encoder $\mathrm{Enc}$; distilled sigma set $\mathcal{S}$; $\lambda_{\text{over}}{=}2.0,\ \lambda_{\text{under}}{=}1.0$}
\tcp{$N$: number of target latent tokens}
\BlankLine
\tcp{(a) multi-operator synthesis}
$T\sim\mathrm{Unif}(\mathcal{T})$;\ \ $S\leftarrow\mathrm{clip}_{[0,1]}(T(H))$;\ \ $o\in[0,1]^{N}$ from $T$\;
\BlankLine
\tcp{(b) flow-matching step (backbone frozen)}
$x_0\leftarrow\mathrm{Enc}(H)$;\quad $c\leftarrow\mathrm{Enc}(S)$;\quad $\epsilon\sim\mathcal{N}(0,I)$;\quad $t\sim\mathcal{U}(\mathcal{S})$\;
$x_t\leftarrow(1-t)x_0+t\epsilon$;\quad $v\leftarrow\epsilon-x_0$;\quad $\delta\leftarrow f_\theta(x_t,t,c)-v\ \in\mathbb{R}^{N\times128}$\;
\BlankLine
\tcp{(c) luminance-supervised weights}
\tcp{acts only where clipping leaves luminance unconstrained}
\For{$n\leftarrow1$ \KwTo $N$}{
 $d_n\leftarrow-\langle \delta_n,w\rangle$ \tcp*{$\propto\langle\hat x_0-x_0,w\rangle$: luminance error}
 \lIf{$d_n>0$}{$\lambda_n\leftarrow\lambda_{\text{over}}$ \tcp*[f]{too bright}}
 \lElse{$\lambda_n\leftarrow\lambda_{\text{under}}$}
 $\text{tok\_w}_n\leftarrow1+(\lambda_n-1)(1-o_n)$ \tcp*{$o_n{=}1\Rightarrow\text{tok\_w}_n{=}1$}
}
$\mathcal{L}(\theta)\leftarrow\frac{1}{N}\sum_{n=1}^{N}\text{tok\_w}_n\lVert \delta_n\rVert_2^{2}$;\quad $\theta\leftarrow\theta-\eta\nabla_\theta\mathcal{L}(\theta)$\;
\caption{One training step.}
\label{alg:train}
\end{algorithm}

\subsection{Multi-operator synthesis}
\S3 showed the between-operator term is removed only by marginalizing $T$. We approximate $\int p(S\mid H,T)\,p(T)\,dT$ by expanding every clip through an explicit family of five operators and sampling uniformly over the result: two grading-style curves, a hard clip, a Reinhard variant \citep{reinhard} and an ACES approximation \citep{aces}. All members but one share the auto-exposure setpoint ($0.18$); all share unit mid-tone slope and differ in shoulder and clipping (Fig.~\ref{fig:arch}, left); parameters are in the supplement. The hard clip is deliberate: identifiability vanishes there (Proposition~\ref{prop:ident}), and a family that omits the extreme leaves the hardest regime untrained. Content combines real captured HDR \citep{hdm2014} with synthetic renders; the mixture addresses the prior factor $p(H)$ while the operator family addresses the likelihood. The corpus is $347$ source clips ($187$ real, the rest synthetic), a source scale comparable to the roughly $300$ clips of the strongest baseline. The five-operator family expands these into $1{,}735$ training pairs. The family varies tone operators only, not signal degradations such as codec noise, which randomized restoration pipelines address instead \citep{realesrgan}; the target is the between-operator term of Eq.~\eqref{eq:lotv}, not generic robustness. The tower property drives the residual mean to zero only over $\mathrm{supp}\,p(T)$ (\S3), so coverage of the support matters more than the number of operators. \S5 measures the reduction this buys and the boundary where the support condition fails.

\begin{table*}[tb]
    \centering\small
    \setlength{\tabcolsep}{4.5pt}
    \begin{tabular}{lccccccc}
    \toprule
    & \multicolumn{4}{c}{seen operators} & \multicolumn{2}{c}{unseen operators} & \\
    \cmidrule(lr){2-5}\cmidrule(lr){6-7}
    Method \hfill {\scriptsize RMSE (bias)} & Hard-Clip & ACES & Reinhard & Grading & Filmic & Severe & mean RMSE $\downarrow$ \\
    \midrule
    \textbf{ReITM (ours)} & $\mathbf{.149}$ {\scriptsize$(+.03)$} & $\mathbf{.153}$ {\scriptsize$(+.02)$} & $\underline{.123}$ {\scriptsize$(+.02)$} & $\underline{.121}$ {\scriptsize$(+.01)$} & $\mathbf{.132}$ {\scriptsize$(+.02)$} & $\underline{.170}$ {\scriptsize$(+.04)$} & $\mathbf{.141}$ \\
    \quad ReITM (w/o temporal pretext) & $\underline{.153}$ {\scriptsize$(+.03)$} & $.166$ {\scriptsize$(+.01)$} & $.129$ {\scriptsize$(+.01)$} & $.127$ {\scriptsize$(+.01)$} & $.137$ {\scriptsize$(+.03)$} & $\mathbf{.169}$ {\scriptsize$(+.04)$} & $.147$ \\
    \quad ReITM (multi-operator only) & $.154$ {\scriptsize$(+.03)$} & $\underline{.160}$ {\scriptsize$(+.02)$} & $.126$ {\scriptsize$(+.02)$} & $.125$ {\scriptsize$(+.02)$} & $\underline{.135}$ {\scriptsize$(+.03)$} & $.173$ {\scriptsize$(+.04)$} & $\underline{.146}$ \\
    \midrule
    LumiVid & $.252$$^{*}$ {\scriptsize$(+.09)$} & $.234$$^{*}$ {\scriptsize$(-.01)$} & $\mathbf{.118}$ {\scriptsize$(+.00)$} & $\mathbf{.114}$ {\scriptsize$(-.00)$} & $.148$ {\scriptsize$(+.01)$} & $.205$$^{*}$ {\scriptsize$(+.09)$} & $.179$ \\
    X2HDR & $.214$$^{*}$ {\scriptsize$(+.08)$} & $.205$$^{*}$ {\scriptsize$(-.04)$} & $.137$ {\scriptsize$(-.01)$} & $.142$$^{*}$ {\scriptsize$(-.01)$} & $.147$ {\scriptsize$(-.01)$} & $.246$$^{*}$ {\scriptsize$(+.12)$} & $.182$ \\
    HDRTVNet++ & $.254$$^{*}$ {\scriptsize$(-.05)$} & $.476$$^{*}$ {\scriptsize$(-.26)$} & $.200$$^{*}$ {\scriptsize$(-.11)$} & $.212$$^{*}$ {\scriptsize$(-.11)$} & $.278$$^{*}$ {\scriptsize$(-.10)$} & $.174$ {\scriptsize$(+.01)$} & $.265$ \\
    LEDiff & $.311$$^{*}$ {\scriptsize$(+.02)$} & $.304$$^{*}$ {\scriptsize$(-.06)$} & $.368$$^{*}$ {\scriptsize$(-.02)$} & $.376$$^{*}$ {\scriptsize$(-.02)$} & $.378$$^{*}$ {\scriptsize$(-.01)$} & $.312$$^{*}$ {\scriptsize$(+.01)$} & $.341$ \\
    \bottomrule
    \end{tabular}
    \caption{Cells: log-luminance error RMSE $L$, lower better (\textbf{bold} best, \underline{underline} second), bias $b$ in parentheses, two decimals ($+$ $=$ too bright). $^{*}$: our lead is significant (bootstrap and Wilcoxon). Seen columns show one of the family's two gradings (\S4.2). Baselines: LumiVid \citep{lumivid}, X2HDR \citep{x2hdr}, HDRTVNet++ \citep{hdrtvnetpp}, LEDiff \citep{lediff}.}
    \label{tab:tilt}
\end{table*}

\subsection{Implementation}
The backbone is a pretrained LTX-2 video-diffusion model \citep{ltx2} with an in-context LoRA of rank $64$ \citep{lora,iclora}, trained for $2{,}500$ steps at batch size $4$; inference is single-stage on the native $8$-step distilled schedule. The SDR conditions the model in context: its latent tokens are concatenated to the noisy target along the sequence axis and the two are attended jointly, so no conditioning module is added. The adapters sit on the attention and feed-forward projections and the backbone weights are never updated, which is how both branches of \S4 keep the pretrained prior. Timesteps are drawn uniformly from the seven interior sigmas $\mathcal{S}$ of that distilled trajectory with $\pm0.02$ jitter rather than from $\mathcal{U}(0,1)$, so training sees the noise levels inference visits; every arm compared in \S5 shares this choice. The full stack, ReITM, adds a temporal in-context pretext task: clipped tokens whose surface is observed in another frame have their noised input replaced by pure noise, so the model has to recover them from cross-frame correspondence alone. The pretext is an implementation detail, not a contribution, and is ablated separately. Operator parameters, the observability formula, and training configuration are in the supplement. Each stack trains in $8.5$ hours on one H100 NVL ($96$\,GB) GPU.

\section{Experiments}
\paragraph{Protocol.}
We evaluate on three benchmarks: \textbf{Hard-Clip} (unseen content, seen hard-clip operator), \textbf{Filmic} (unseen soft operator, the Hable curve \citep{hable2010}) and \textbf{Severe} (unseen: setpoint $0.35$ and mid-tone $\gamma=0.6$, both outside the training family), on $84$ clips ($42$ ARRI, $30$ UPIQ \citep{upiq} and $12$ LiU HDRv \citep{kronander2014}; the UPIQ references are photometrically calibrated still images, each rendered as a static nine-frame clip following \citet{lumivid}). The primary metric is the luminance error $L$ with its bias $b$ (\S3), corroborated by CVVDP JOD at 1000-nit BT.709 \citep{cvvdp} and PU21-PSNR \citep{pu21}; $L$, $b$ and PU21-PSNR use the median-luminance alignment of \S3 and CVVDP the automatic per-clip policy, both levels swept in Table~\ref{tab:aligned}, with no re-fitted response curve. Baselines run through their own released code paths, single-stage (details: supplement). Within generative video ITM this is, to our knowledge, the first evaluation on tone operators fully excluded from training.

\begin{figure}[b]
\centering
\includegraphics[width=0.94\linewidth]{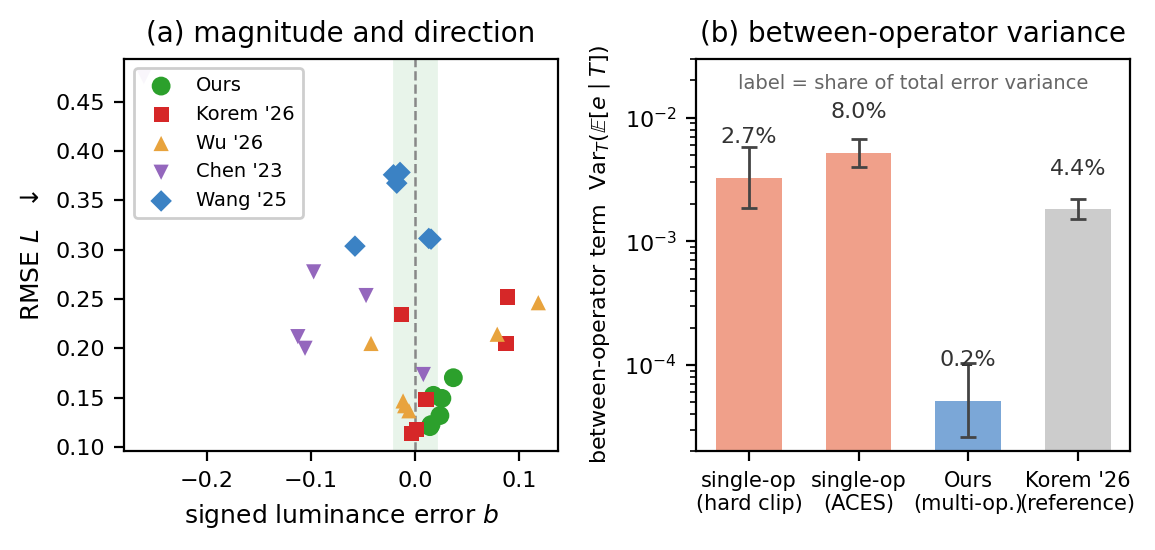}
\caption{\textbf{(a)} One point per method and operator. Low $L$ and $b{\approx}0$ is better. \textbf{(b)} Between-operator term of Eq.~\eqref{eq:lotv}, log scale.}
\label{fig:tilt}
\end{figure}

\paragraph{Luminance error.}
ReITM attains the lowest RMSE of any public method on all three benchmarks and the lowest mean over the six-operator evaluation grid ($0.141$ versus $0.179$--$0.341$). Across the grid we are ahead in $22$ of the $24$ method-by-operator cells and significantly ahead in $18$ (asterisks in Table~\ref{tab:tilt}), and no baseline lead has an interval excluding zero. On Filmic the margin is significant against \citet{hdrtvnetpp} and \citet{lediff}, and not against \citet{lumivid} or \citet{x2hdr}. \citet{lumivid} leads only on Reinhard and Grading (Table~\ref{tab:tilt}), both soft operators. The bias measures direction: ReITM stays within $+0.01$ to $+0.04$ on every operator while the baselines swing with it (Fig.~\ref{fig:tilt}a). Fig.~\ref{fig:concept}b shows the same reversal on one clip. The two components act on different parts: multi-operator training controls the magnitude, and the asymmetric loss the bias (ACES $+.025\!\to\!+.013$ against multi-operator training). Stopping down the reconstructions separates recovered structure from a bright constant fill (Fig.~\ref{fig:teaser}).

\begin{table}[hbt!]
    \centering\footnotesize
    \setlength{\tabcolsep}{3.5pt}
    \begin{tabular}{lccc}
    \toprule
    Method & Hard-Clip & Filmic & Severe \\
    \midrule
    \textbf{ReITM (ours)} & \textbf{7.69/28.07} & \textbf{8.57/31.69} & \textbf{6.88/27.66} \\
    LumiVid & 6.12/21.67 & 8.38/31.20 & 6.16/23.45 \\
    HDRTVNet++ & 6.03/24.95 & 7.40/29.06 & 5.59/26.82 \\
    X2HDR & 5.84/23.27 & 7.69/30.68 & 5.37/22.36 \\
    LEDiff & 5.03/23.36 & 3.26/20.99 & 4.79/23.23 \\
    \bottomrule
    \end{tabular}
    \caption{Cells: CVVDP (JOD $\uparrow$, auto) / PU21-PSNR (dB $\uparrow$, median-aligned). Native regime per method (Table~\ref{tab:aligned}).}
    \label{tab:cvvdp}
\end{table}

\paragraph{Controlling direction coincides with the best quality.}
With the direction controlled, the same system leads every benchmark on perceptual quality and fidelity (Table~\ref{tab:cvvdp}). Against LumiVid, second best everywhere, clip-paired tests on CVVDP give $\Delta{=}{+}1.574$ on Hard-Clip ($p{=}3.4{\times}10^{-13}$), $+0.718$ on Severe ($p{=}9.5{\times}10^{-9}$) and $+0.183$ on Filmic ($p{=}9.3{\times}10^{-3}$); we take $75$ of $84$ clips on Hard-Clip and $69$ of $84$ on Severe, counts that move by at most three across the three alignment levels of Table~\ref{tab:aligned}. The Filmic margin is narrow and we report it as remaining ahead, not dominating. PU21-PSNR agrees on all three, so the win is not specific to one perceptual model. The gain concentrates where the observation saturates: restricting PU21-PSNR to SDR luma $\ge0.96$ raises the margin over LumiVid.

\begin{table}[t]
\centering\footnotesize
\setlength{\tabcolsep}{2.5pt}
\begin{tabular}{lccc}
\toprule
Method & Hard-Clip & Filmic & Severe \\
\midrule
\textbf{ReITM (ours)} & \textbf{8.63/6.35/7.76} & \textbf{9.28/7.57/8.57} & \textbf{7.98/5.06/7.54} \\
LumiVid & 6.72/5.01/6.78 & 9.21/7.31/8.17 & 6.94/4.84/6.75 \\
HDRTVNet++ & 7.30/4.64/5.06 & 8.19/6.24/7.55 & 7.52/2.74/5.93 \\
X2HDR & 7.06/5.14/3.34 & 8.98/7.00/4.90 & 6.78/4.50/2.62 \\
LEDiff & 7.05/2.01/5.48 & 5.47/0.21/3.15 & 6.97/1.58/5.16 \\
\bottomrule
\end{tabular}
\caption{Table~\ref{tab:cvvdp} by source. Cells: CVVDP (JOD $\uparrow$), ARRI / UPIQ / LiU. No source drives the lead.}
\label{tab:persource}
\end{table}

\paragraph{Robustness across sources.}
Split by data source (Table~\ref{tab:persource}), our lead holds in all nine cells, generally largest where the analysis predicts the problem is hardest (Hard-Clip and Severe) and small but positive on Filmic.

\paragraph{The full operator grid.}
Extending the CVVDP comparison to all six operators (supplement), we lead on every one except Grading ($8.749$ versus our $8.581$), the soft profile where Table~\ref{tab:tilt} also shows the baseline ahead: robustness costs nothing on any other operator.

\paragraph{The variance diagnosis.}
Eq.~\eqref{eq:lotv} predicts that training on a single operator leaves the between-operator term standing, and we test that prediction directly. Two arms are trained on one operator each, a hard clip and the ACES approximation. The three arms share backbone, source clips, optimization steps and loss, and differ only in how many operators the family contains, which is also what changes the number of training pairs. Decomposing the luminance error over the six-operator grid, the between-operator term is $0.0033$ and $0.0052$ for the two single-operator arms against $0.00005$ under marginalization, that is $2.7\%$ and $8.0\%$ of the total error variance against $0.2\%$ (clip-paired bootstrap, $84$ clips, $2000$ resamples, intervals disjoint; Fig.~\ref{fig:tilt}b). In the primary metric the mean $L$ falls from $0.217$ and $0.224$ to $0.146$, and the operator-to-operator spread from $0.164$ and $0.160$ to $0.048$. Marginalization both lowers the error and flattens it across operators. We make no claim about which single operator is the better choice, since the two arms are within $0.007$ of each other in mean $L$. \citet{lumivid} sits between the two regimes at $4.4\%$; its training operator is not disclosed and we do not attribute this to an operator count.

\paragraph{Ablation.}
Table~\ref{tab:tilt} reads as an ablation: against multi-operator training alone the asymmetric loss reduces the bias on all six operators, three of them below the rounding, and the temporal pretext gives the lowest RMSE overall ($.141$). On CVVDP the staircase is monotone but small and its clip-paired intervals mostly cross zero (only the temporal pretext on Filmic is significant, $\Delta{=}{+}0.046$, $[{+}0.010,{+}0.081]$), so we treat it as descriptive and rest the asymmetric-loss claim on the bias effect. A pre-registered perceptual-study protocol is in the supplement.

\begin{table}[t]
 \centering\footnotesize
 \setlength{\tabcolsep}{2.5pt}
 \begin{tabular}{lccc}
 \toprule
 Method & Hard-Clip & Filmic & Severe \\
 \midrule
 \textbf{ReITM (ours)} & \textbf{6.98/7.69/7.81} & \textbf{7.61/8.57/8.68} & \textbf{6.43/6.88/6.99} \\
 LumiVid & 5.27/6.12/6.53 & 7.34/8.38/8.58 & 5.71/6.16/6.50 \\
 HDRTVNet++ & 5.93/6.03/6.68 & 7.18/7.40/7.85 & 5.82/5.59/5.98 \\
 \bottomrule
 \end{tabular}
 \caption{Each cell: CVVDP under none/auto/median photometric alignment. Auto resolves per clip by whether the reference carries an absolute-nit convention.}
 \label{tab:aligned}
\end{table}

\paragraph{Sensitivity to photometric alignment.}
Table~\ref{tab:aligned} sweeps three photometric-alignment levels under otherwise identical scoring: none, the automatic policy of the primary protocol, and full median-luminance. ReITM is first on every benchmark at every level, nine of nine cases. Alignment helps the strongest baseline more than us: on Hard-Clip it lifts \citet{lumivid} by $+1.26$ JOD against our $+0.83$, narrowing our margin from $1.71$ to $1.28$ without reversing. All three levels are exposure-type, so the sweep measures what a scale match can absorb.

\begin{figure}[hbt!]
    \centering
    \includegraphics[width=\linewidth]{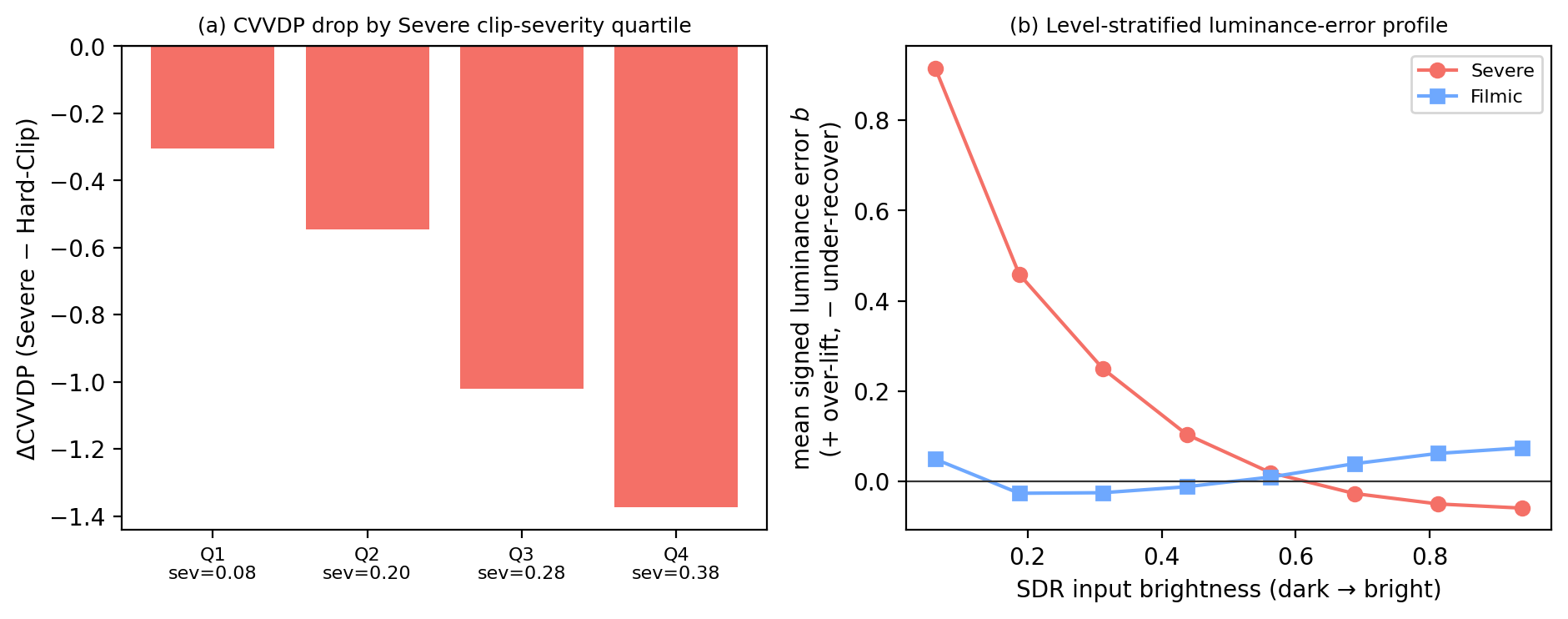}
    \caption{\textbf{(a)}~CVVDP shortfall on Severe relative to Hard-Clip, by clipped-fraction quartile. \textbf{(b)}~Signed luminance bias $b$ by SDR level; Severe (red) and in-support Filmic (blue). Residual failure outside the training operator family; multi-operator-only stack matches.}
    \label{fig:boundary}
\end{figure}

\paragraph{The boundary of marginalization.}
We still lead on Severe (Table~\ref{tab:cvvdp}), yet \S3's range condition predicts residual failure outside $\mathrm{supp}\,p(T)$, and Fig.~\ref{fig:boundary} tests that prediction on our own stack. Against Hard-Clip, the CVVDP shortfall grows with clip severity (correlation $-0.459$) and localizes on the two out-of-family axes: shadow PSNR falls $3.84$ dB versus $0.83$ dB in the highlights, and signed bias decreases monotonically from $+0.92$ in the shadows to $-0.06$ in the highlights. The profile is flat on Filmic ($\pm0.09$) and unchanged for the multi-operator-only arm, so the boundary is imposed by the training family, not the asymmetric loss.

\section{Conclusion}
We cast SDR-to-HDR video reconstruction as blind posterior inference under an unknown tone operator. The framing yields one diagnosis for two symptoms: the operator dependence is the between-operator variance term, removable by training-time marginalization; the directional bias is a misspecified action along the latent luminance coordinate, corrected by an observability-gated asymmetric loss whose optimal action the analysis identifies. Controlling that direction coincides with the best quality on all three benchmarks, and the same decomposition localizes our residual to the axes the training family holds fixed. Two limitations bound the claim: marginalization holds only over the training family's span (Fig.~\ref{fig:boundary}), and each stack is one training run.

\clearpage
\bibliography{refs}

\clearpage
\appendix
\setcounter{proposition}{0}
\numberwithin{table}{section}
\numberwithin{figure}{section}
\numberwithin{equation}{section}
\setcounter{topnumber}{3}
\setcounter{bottomnumber}{2}
\setcounter{totalnumber}{5}
\setcounter{dbltopnumber}{3}
\renewcommand{\topfraction}{0.92}
\renewcommand{\bottomfraction}{0.5}
\renewcommand{\textfraction}{0.06}
\renewcommand{\floatpagefraction}{0.80}
\renewcommand{\dbltopfraction}{0.92}
\renewcommand{\dblfloatpagefraction}{0.80}

\twocolumn[\begin{center}{\LARGE\bfseries Supplementary Material\par}\end{center}\vskip 2em]

\noindent
This supplement accompanies the main paper. Section references of the form
\S3 or \S4.1 point to the main paper; references to Appendix~A and later
point to this supplement. The main paper is self-contained; nothing here is
required to follow its argument.

\paragraph{Organization}
Appendix~\ref{app:ident} proves the properties stated in \S3.
Appendix~\ref{app:probe} covers the latent luminance readout and the scope of the fit.
Appendix~\ref{app:asym} proves the properties behind the asymmetric loss and the observability gate.
Appendix~\ref{app:impl} gives operator parameters, the observability formula and the training configuration.
Appendix~\ref{app:baselines} gives the baseline inference regimes.
Appendix~\ref{app:grid} gives the full operator evaluation grid.
Appendix~\ref{app:singleop} reports the single-operator control arms.
Appendix~\ref{app:moved} expands tables from the main text.
Appendix~\ref{app:qual} extends the qualitative comparison.
Appendix~\ref{app:limits} states the limitations in full and reports a deployment-time operator-inference attempt that did not work.
Appendix~\ref{app:repro} summarizes reproducibility information.

\input{supp/A_identifiability}
\input{supp/B_probe}
\input{supp/C_asym}
\input{supp/D_implementation}
\input{supp/E_baselines}
\input{supp/F_grid}
\input{supp/G_singleop}
\input{supp/H_moved}

\input{supp/I_qualitative}
\input{supp/L_limitations}
\input{supp/M_reproducibility}

\end{document}

%% file: supp/A_identifiability.tex
\section{Identifiability, Marginalization, and the Range-Invariance of the Luminance Bias}
\label{app:ident}

This appendix gives proof sketches for the results stated in \S3 and for the
range-invariance of the signed luminance bias. Standing assumptions are listed once. We work per token on the log-luminance coordinate. The scene log-luminance is $X \sim p(X)$. The observation is $Y = \min(g_\theta(X), 1)$, where $g_\theta$ is a monotone increasing tone curve indexed by $\theta \sim \pi$ with $\operatorname{supp}\pi = \Theta_{\mathrm{train}}$. The observation is treated as noiseless. These simplifications (scalarization to the luminance coordinate, deterministic $g$, monotonicity) are used throughout Appendices~\ref{app:ident} and~\ref{app:asym}.

\subsection{Partial identifiability}

\begin{proposition}[Partial identifiability]
\label{prop:ident-app}
On the un-saturated region the pair (tone curve $f$, exposure $e$) acts as an invertible measurement, so the scene radiance is identifiable conditional on the operator; on the clipped region neither the operator nor the radiance is identifiable, and identifiability degrades continuously as the clipped fraction grows.
\end{proposition}

\begin{proof}[Proof sketch]
For a token with $Y < 1$, the observation equation $Y = g_\theta(X)$ with $g_\theta$ strictly monotone gives $X = g_\theta^{-1}(Y)$: conditional on $\theta$, the likelihood is a point mass and $X$ is exactly determined. Across $\theta$ the candidate set $\{g_\theta^{-1}(Y) : \theta \in \operatorname{supp}\pi\}$ is bounded by the envelope of the inverse family, so uncertainty in $X$ is bounded by the spread of the family at that observation level. For a token with $Y = 1$, the likelihood is constant on the event $\{X \ge g_\theta^{-1}(1)\}$: every $X$ above the saturation threshold produces the identical observation, so the likelihood carries no information beyond a one-sided constraint, and the posterior on the clipped region equals the prior truncated from below. The monotonicity argument for continuous degradation is as follows. Let $c_\theta = g_\theta^{-1}(1)$ denote the saturation threshold in scene units. As $c_\theta$ decreases (more aggressive clipping), the event $\{X \ge c_\theta\}$ grows monotonically, the mass of the observation carried by the constant-likelihood branch increases monotonically, and the identified region shrinks monotonically. Identifiability is therefore not a binary property of a frame but a per-token quantity that decays monotonically with the clip threshold; this is what the observability variable $o \in [0,1]$ tracks, with $o \to 1$ for unclipped tokens and $o \to 0$ for saturated tokens.
\end{proof}

\subsection{Marginalization effect and its range}

\begin{lemma}[Marginalization effect and its range]
\label{lem:marg}
Training across an operator family removes the between-operator variance component of the error, but only over $\operatorname{supp} p(T)$.
\end{lemma}

\begin{proof}[Proof sketch]
Write the reconstruction error $\varepsilon$ and expand by the law of total variance with respect to the operator $T$:
\[
\begin{aligned}
\mathrm{Var}(\varepsilon) &= \mathbb{E}_T\big[\, \mathrm{Var}(\varepsilon \mid T) \,\big] + \mathrm{Var}_T\big(\, \mathbb{E}[\varepsilon \mid T] \,\big)\\
&= \text{within-operator} + \text{between-operator}.
\end{aligned}
\]
An estimator trained against a single operator $T_0$ is a function calibrated to the conditional law $p(S \mid H, T_0)$; evaluated under $T \neq T_0$ its conditional mean error $\mathbb{E}[\varepsilon \mid T]$ varies with $T$, and the second term is strictly positive whenever the family is non-degenerate. An estimator trained with $T$ drawn from $p(T)$ targets the marginal posterior $p(H \mid S) \propto p(H) \int p(S \mid H, T)\, p(T)\, dT$; its conditional error given the observation is centered by construction (tower property, developed as Proposition~\ref{prop:misspec}(a) in Appendix~\ref{app:asym}). Centering holds level-wise in the observation and drives the family-averaged conditional error to zero; it does not force $\mathbb{E}[\varepsilon \mid T]$ to vanish for each individual $T$, so the between-operator term is driven down rather than annihilated. Appendix~\ref{app:singleop} measures the residual at $0.16\%$ of the total error variance, against $2.7\%$ and $8.0\%$ for the single-operator arms. The removal argument uses only that the test operator lies in $\operatorname{supp} p(T)$: every step conditions on $T \in \operatorname{supp}\pi$. For a test operator outside the support the marginal posterior assigns it zero prior weight, so no cancellation is guaranteed; whatever transfer occurs reflects how close the test curve sits to the family rather than the argument; this is the range condition, and it is the theoretical preview of the boundary analysis in \S5.
\end{proof}

\subsection{Exposure-affine luminance readout}

\begin{proposition}[Exposure-affine luminance readout]
\label{prop:probe-app}
An exposure change acts to first order as a translation of the latent along a fixed direction $u$. Consequently, for any fixed linear readout $r(z) = \langle z, w\rangle + \beta$, the readout is affine in log-exposure with a single gain $g = \langle u, w\rangle$, and log-luminance is recoverable from $r$ up to that affine calibration. The readout direction $w$ need not coincide with $u$; only their inner product enters.
\end{proposition}

\begin{proof}[Proof sketch]
An exposure change multiplies linear-domain luminance by a constant $a > 0$, which is the additive shift $\log a$ on the log-luminance axis. If the encoder restricted to the luminance factor is differentiable in log-exposure, a first-order expansion around the operating point gives $z(a) \approx z(1) + (\log a) \cdot u$ with $u$ the directional derivative of the latent with respect to log-exposure; exposure sweeps then trace straight lines in latent space, and any linear functional $\langle \cdot, w \rangle$ with $\langle u, w \rangle \ne 0$ recovers log-luminance up to affine calibration. Appendix~\ref{app:probe} proves the affine form and shows that $w$ is not required to be parallel to $u$. The proposition is an approximation claim, and its support is empirical: a fixed linear probe on the LTX-2 VAE latent attains held-out $R^2 \approx 0.9$ for log-luminance, extends to per-channel ($\approx 0.84$--$0.96$) and color-opponent ($\approx 0.88$) axes. A PCA control on 16 held-out clips (57{,}120 tokens) shows the direction is not the latent's dominant mode: it lies $82.3^\circ$ from PC1, no single principal component exceeds $R^2 = 0.275$ for log-luminance (PC1 $= 0.155$), reaching the probe's accuracy from principal components alone takes about 50 of them, and 20 random unit directions give mean $R^2 = 0.079$, so the readout must be fitted, not recovered from the leading variance structure. On a second backbone (Wan2.1-VACE-14B, exposure sweep EV $\in \{-2 \ldots 2\}$, leave-one-clip-out over 8 clips) the probe attains $R^2 = 0.592$, with 7/8 clips clearly positive and one clip negative ($-0.365$, narrow exposure range). We therefore scope Proposition~\ref{prop:probe} as demonstrated on LTX-2 with partial cross-backbone support, not as backbone-universal.
\end{proof}

\subsection{Range-invariance of the signed luminance bias}

The signed luminance bias $b$ is the mean of (prediction $-$ GT) along the log-luminance axis; positive $b$ is over-expansion. The derivation of range-invariance proceeds through the exposure quotient. Let both prediction and ground truth be rescaled by a common exposure factor $a > 0$, as happens when the same delivery convention is applied to both. The alignment protocol of \S3 instead fits one factor to the prediction alone; that operation changes $b$ by construction, which is why the aligned $b$ is reported as the quantity of interest rather than treated as invariant to it. On the log axis the rescaling adds the same constant $\log a$ to $\hat{X}$ and to $X$, so the difference is unchanged:
\[
\begin{aligned}
b(a\hat{H}, aH) &= \mathbb{E}\big[\, (\hat{X} + \log a) - (X + \log a) \,\big]\\
&= \mathbb{E}\big[\, \hat{X} - X \,\big] \;=\; b(\hat{H}, H).
\end{aligned}
\]
Equivalently, $b$ depends on the two signals only through the quotient $\hat{H}/H$, which is invariant to common rescaling. It is therefore insensitive to the overall output range convention of a method and measures directional bias only; it is orthogonal to magnitude metrics. This invariance also composes with property (i) stated alongside Proposition~\ref{prop:expectile} in Appendix~\ref{app:asym}: the expectile action is location--scale equivariant, so the asymmetric intervention commutes with exposure scaling and does not interact with the alignment protocol.

%% file: supp/B_probe.tex
\section{The Latent Luminance Readout: Proof and Limits}
\label{app:probe}

\paragraph{Statement}
Let $E$ be the frozen video autoencoder, $z = E(H)$, and $s = \log_{10}\alpha$ the
log-exposure. Let $r(z) = \langle z, w\rangle + \beta$ be the fitted readout with intercept $\beta$
(the symbol $b$ is reserved for the signed luminance bias elsewhere), let
$u = \partial E/\partial s|_{s=0}$ be the exposure displacement, and let
$g = \langle u, w\rangle$. Proposition~\ref{prop:probe} states that for any fixed $(w,\beta)$,
\begin{equation}
r\!\left(E(10^{s}H)\right) = r\!\left(E(H)\right) + g\cdot s + O(s^{2}).
\label{eq:probe-affine}
\end{equation}
Three consequences follow, and the proof below establishes them in turn.
(i) Only the component of the exposure displacement along $w$ is read; everything
orthogonal to $w$ is annihilated.
(ii) Unit gain $g=1$ is a single scalar constraint on the product
$\lVert u\rVert\,\lVert w\rVert\cos\angle(u,w)$, so it does not require $w \parallel u$.
(iii) The intercept cancels whenever the readout is applied to a difference, so no
calibration of $b$ is needed for the directional signal used in \S4.1.

\begin{proof}
Write $z(s) := E(10^{s}H)$, a curve through $z(0) = E(H)$. Since $r$ is affine,
$\tfrac{d}{ds} r(z(s)) = \langle z'(s), w\rangle$, and with $z'$ bounded on the working
range,
\[
\begin{aligned}
r(z(s)) - r(z(0)) &= \int_{0}^{s} \langle z'(\sigma), w\rangle \, d\sigma\\
&= g\cdot s + O(s^{2}), \qquad g = \langle z'(0), w\rangle .
\end{aligned}
\]
Only the scalar $g$ survives: decomposing
$d_{s} := z(s) - z(0) = \left(\langle d_{s}, w\rangle / \|w\|^{2}\right) w + d_{s}^{\perp}$
with $\langle d_{s}^{\perp}, w\rangle = 0$, the off-$w$ component is annihilated by $r$,
which is consequence (i).
For (ii), $g = \|z'(0)\| \cdot \|w\| \cdot \cos\angle(z'(0), w)$, so $g = 1$ is one scalar
constraint on a product of three factors. Any $w$ with $\cos \neq 0$ can be scaled to
satisfy it, the solution set is the affine hyperplane $\{w : \langle u, w\rangle = 1\}$ of
dimension $\dim z - 1$, and $\angle = 90^\circ$ is the only failure mode. Parallelism is
sufficient and unnecessary.
For (iii), the intercept cancels in differences with no linearization. In the
flow-matching parameterization of Algorithm 1 the implied clean-latent error is
$\hat{x}_{0} - x_{0} = -t\cdot e$ with $t \in (0,1)$, so
$-\langle e, w\rangle = t^{-1}\langle \hat{x}_{0} - x_{0}, w\rangle$ is a strictly positive
multiple of the readout error. The loss thresholds that scalar at zero, so it is invariant
to $t$, to $\|w\|$, and to any positive gain error.
\end{proof}

\paragraph{The gain is a pairing, not an alignment}
This observation is moved here from the main text. The exposure displacement is a
translation. Within a clip it is linear in $s$ and its normalized increments are mutually
collinear (cosine $0.84$--$0.99$). It does not, however, travel along $w$: the measured $u$
makes $46^\circ$--$59^\circ$ with $w$, so only $27$--$48\%$ of $\|u\|^{2}$ is visible to the
readout. Near-unit gain and a large angle coexist because
$g = \|u\| \cdot \|w\| \cdot \cos\angle(u, w)$. Alignment is the wrong target. Projecting on
the exposure direction instead gives the \emph{worse} luminance readout (per-token $R^{2}$
$0.06$ / $0.59$ / $0.80$ against $0.82$ / $0.82$ / $0.90$), and $u$ is only approximately
clip-independent (cross-clip cosine $0.68$--$0.94$) while $w$ is fixed by construction. We
use the fitted readout throughout and claim nothing about the geometry of the displacement.

\paragraph{Analytic input and empirical bounds}
The only analytic input is smoothness of $E$ in $s$. The rest is bounded empirically. Over
$s \in [-0.6, +0.6]$ the readout is affine in $s$ with maximum residual
$0.013$ / $0.005$ / $0.040$ decades ($R^{2} = 0.9996$ / $0.99995$ / $0.9953$), the largest
belonging to the clip with the widest dynamic range. The values of $(w, b)$ are not derived.
They are a ridge least-squares fit with per-channel standardization folded into
$w_{\mathrm{eff}} = w/\sigma$ and into $b$, validated on held-out clips. Near-unit gain is
therefore partly a property of that fit: a probe regressed against $\log_{10} Y$ inherits
unit slope wherever the encoder is exposure-equivariant. We present $g \approx 1$ as a
verified consistency of two calibrations, not as an independent discovery.

\paragraph{Per-clip sweep}
Table~\ref{tab:probe-sweep} reports the per-clip sweep.

\begin{table*}[t]
\centering
\scriptsize
\setlength{\tabcolsep}{4pt}
\begin{tabular}{lccccccc}
\toprule
clip & gain $g$ & $R^{2}$ & $g$ vs true & $|\cos\angle(u,w)|$ & $\|d_{s}\|^{2}$ & per-token & per-token \\
 & (readout vs $s$) & & $\log Y$ & & along $w$ & $R^{2}$: $w$ & $R^{2}$: $u$ \\
\midrule
\texttt{arri\_M001C001\_161207\_R00H} & 1.068 & 0.9996 & 1.068 & 0.68--0.70 & 47--49\% & 0.817 & 0.064 \\
\texttt{liu\_bridge\_w0} & 0.980 & 0.99995 & 1.024 & 0.64--0.65 & 41--43\% & 0.816 & 0.593 \\
\texttt{hdm\_smith\_welding\_0120}\,$\dagger$ & 0.905 & 0.9953 & 0.931 & 0.46--0.55 & 21--30\% & 0.902 & 0.799 \\
\bottomrule
\end{tabular}
\caption{Per-clip latent exposure sweep. $\dagger$ scene shares content with the probe fit
set. Cross-clip $\cos(u_{i}, u_{j}) = 0.939$ (arri--liu), $0.835$ (liu--hdm), $0.679$
(arri--hdm): the exposure direction is only approximately content-independent. UPIQ was
excluded from the sweep because its EXRs are in absolute cd/m$^{2}$, so $59$--$72\%$ of
pixels saturate the LogC3 encoding and a sweep there would measure clipping rather than
translation.}
\label{tab:probe-sweep}
\end{table*}

\paragraph{Limits of the readout}
These limits are stated here rather than in the main text.

\begin{description}
\item[Range and scope.] The sweep covers three clips over $\pm 2$ EV on one backbone. The
gain is $0.905$--$1.068$ there and we claim no linearity outside that range. One of the
three clips shares a scene with the fit set; it is marked in Table~\ref{tab:probe-sweep}
and is not treated as held out.

\item[Clipped tokens.] Unit gain and readout accuracy are measured on unclipped latents,
while the gate applies the asymmetry precisely where clipping dominates. Accuracy on
clipped tokens is inferred from the end-to-end bias reduction, not measured directly.

\item[Per token vs per clip.] Near-unit gain is a clip-level property of the token-mean
readout. Per token the displacement is heterogeneous (mean $\cos(d_{s}, w) = 0.27$--$0.64$),
and the per-token licence comes from the regression fit ($R^{2} = 0.82$--$0.90$), not from
the translation result.

\item[Read map, not write map.] Displacing a latent by $c \cdot w/\|w\|^{2}$ moves decoded
log-luminance by only $\approx 0.19c$, though uniformly (IQR $0.01$) and
structure-preservingly (correlation $0.997$), on one clip. Our loss performs no such write.
The number bounds any future use of $w$ as an edit direction.

\item[Cross-backbone.] A single alternative autoencoder, where a probe fitted the same way
reaches $R^{2} = 0.592$ with one held-out clip negative, over an eight-clip
leave-one-out exposure sweep. We did not repeat the gain and angle measurements of
Table~\ref{tab:probe-sweep} on it.

\item[Origin of the log-linearity.] It is partly inherited from the log-domain
pre-encoding of the backbone's input. The claim is that a single global affine functional
survives a nonlinear encoder with $8 \times 8 \times 8$ spatiotemporal downsampling trained
under no luminance objective. We do not claim it for linear-light or display-gamma inputs.
\end{description}

%% file: supp/C_asym.tex
\section{Proofs for the Asymmetric Loss and the Observability Gate}
\label{app:asym}

Notation continues from Appendix~\ref{app:ident}, with one addition local to this
appendix. The luminance error of a prediction is written $\varepsilon_{\mathrm{lum}} = \hat{X} - X$,
read through the probe coordinate as
$\varepsilon_{\mathrm{lum}} = \langle \hat{z} - z, w \rangle$. The symbol $u$ keeps the meaning it has
in Appendices~\ref{app:ident} and~\ref{app:probe}, the exposure displacement direction. In the implementation the sign is computed as
\texttt{lum\_dir} $= -\langle e, w_{\mathrm{eff}} \rangle$, where $w_{\mathrm{eff}}$
is the probe direction $w$ expressed in the trainer's raw latent basis and $e$ is
the velocity error, so \texttt{lum\_dir} $> 0$ indicates a prediction brighter than
the ground truth. The asymmetric weights are $\lambda_{\mathrm{over}} = 2.0$ and
$\lambda_{\mathrm{under}} = 1.0$, and the token gate is
\texttt{tok\_w} $= 1 + (\lambda - 1)(1 - o)$. We state up front that
$\lambda = 2/1$ is a tuned value: the theory below predicts the sign of the useful
asymmetry and the existence of an interior optimum, and the magnitude is selected
on validation data (``sign-predicted, magnitude-validated'').

\subsection{Asymmetric quadratic loss selects an expectile action}

\begin{proposition}[Expectile action]
\label{prop:expectile}
For the loss
$\ell(a; x) = \lambda_{\mathrm{over}} (a-x)^2 \mathbf{1}(a > x)
+ \lambda_{\mathrm{under}} (x-a)^2 \mathbf{1}(a \le x)$,
the minimizer of the conditional risk $\mathbb{E}[\ell(f(S); X) \mid S]$ exists,
is unique for every $S$, and equals the $\tau$-expectile of the posterior,
\[
f^*(S) = e_\tau\big( p(X \mid S) \big),
\qquad
\tau = \frac{\lambda_{\mathrm{under}}}{\lambda_{\mathrm{over}} + \lambda_{\mathrm{under}}}.
\]
\end{proposition}

With $\lambda = 2/1$, $\tau = 1/3 < 1/2$, so the action moves darker than the
posterior mean. This is the asymmetric least squares functional of
\citep{newey1987}.

\begin{proof}[Proof (first-order condition)]
The conditional risk
$R(a) = \lambda_{\mathrm{over}} \mathbb{E}[(a-X)^2 \mathbf{1}(X < a) \mid S]
+ \lambda_{\mathrm{under}} \mathbb{E}[(X-a)^2 \mathbf{1}(X \ge a) \mid S]$
is differentiable and strongly convex in $a$, with
\[
\begin{aligned}
&R'(a) = 2\lambda_{\mathrm{over}} \mathbb{E}[(a-X)^+ \mid S]
- 2\lambda_{\mathrm{under}} \mathbb{E}[(X-a)^+ \mid S] = 0\\
&\iff\;
\tau \, \mathbb{E}[(X-a)^+ \mid S] = (1-\tau) \, \mathbb{E}[(a-X)^+ \mid S].
\end{aligned}
\]
The left side is monotone decreasing in $a$ and the right side monotone
increasing, so the root is unique; this identity is the defining equation of the
$\tau$-expectile. At $\tau = 1/2$ the condition reads
$\mathbb{E}[(X-a)^+] = \mathbb{E}[(a-X)^+]$, which holds iff
$a = \mathbb{E}[X \mid S]$, recovering the mean.
\end{proof}

Three companion properties are stated with the proposition; each has a
two-to-three-line proof.

\begin{itemize}
\item \textit{(i) Location--scale equivariance.} $e_\tau(aX + b) = a \cdot e_\tau(X) + b$
for $a > 0$, by substitution in the first-order condition. Consequence: the
expectile intervention commutes with exposure scaling and is consistent with the
range-invariant design of the signed luminance bias (Appendix~\ref{app:ident}) and with the alignment
protocol of \S5.
\item \textit{(ii) Monotonicity in $\tau$.} $e_\tau$ is monotone increasing in $\tau$;
for symmetric distributions $e_\tau + e_{1-\tau} = 2\mu$. Consequence: the
$\lambda$ ratio is a continuous bias dial, used in the existence argument of
Proposition~\ref{prop:gate}(ii).
\item \textit{(iii) Elicitability.} The asymmetric quadratic loss is a strictly
consistent scoring function for the expectile, so the target the network
approximates under finite samples and SGD is well defined. Among generalized
quantiles, expectiles are the only coherent risk measures \citep{bellini2014},
and they are the unique coherent risk measures that are also elicitable
\citep{ziegel2016}; a
deep-learning precedent for expectile regression with the same stability
motivation is IQL \citep{iql2022}.
\end{itemize}

For a standard normal posterior the closed form gives
$e_{1/3} \approx \mu - 0.28\sigma$ (a toy-Gaussian anchor, not a claim about the
learned posterior): $\lambda = 2/1$ asks for an action about $0.28$ posterior
standard deviations darker than the mean, and because of the $(1 - o)$ gate the
request is active only where the posterior uncertainty is actually large, so the
absolute shift self-scales with $\sigma$.

\begin{remark}[Gap to the implemented loss]
\label{rem:asym-gap}
The implementation weights the whole token residual $\|e\|^2$ by the sign of $\varepsilon_{\mathrm{lum}}$
rather than weighting the luminance coordinate alone. Decomposing
$e = \varepsilon_{\mathrm{lum}} \cdot w + e_\perp$: the luminance component receives exactly the asymmetric least
squares above and converges to the expectile; the orthogonal component $e_\perp$
receives a weight depending on $\mathrm{sign}(\varepsilon_{\mathrm{lum}})$, but the coupling between
$\mathrm{sign}(\varepsilon_{\mathrm{lum}})$ and the direction of $e_\perp$ is a second-order effect and the
expected weight on $e_\perp$ is sign-symmetric, so coordinates orthogonal to $w$
are unbiased to first order. The projection-based reading of the loss is therefore
a first-order approximation, stated explicitly here.
\end{remark}

\begin{remark}[Why expectile rather than quantile]
\label{rem:asym-quantile}
The pinball loss has a piecewise-constant gradient, which conflicts with the L2
scale of the $x_0$-prediction diffusion objective; the expectile loss keeps the L2
gradient structure and changes only the weights, which is precisely the two-line
implementation used. Expectiles also use the magnitude information of both tails,
improving sample efficiency; quantile regression is \citep{koenker1978}, and the
LINEX asymmetric Bayes line is \citet{zellner1986}. Clipping itself is Tobit-type
censoring \citep{tobin1958}.
\end{remark}

\subsection{Misspecification determines the zero, sign, and location of the bias}

\begin{proposition}[Misspecification and the direction of the bias]
\label{prop:misspec}
Let $\hat{X}_\pi(y) = \mathbb{E}_\pi[X \mid Y = y]$ be the posterior-mean estimator
under the training prior $\pi$.
(a) If the data are generated with $\theta \sim \pi$, then
$b = \mathbb{E}[\hat{X}_\pi(Y) - X] = 0$, and moreover
$\mathbb{E}[X - \hat{X}_\pi(y) \mid Y = y] = 0$ at every observation level.
(b) If the data are generated by a fixed $\theta^*$ with
$g_{\theta^*} \ge g_\theta$ for all $\theta \in \mathrm{supp}\,\pi$ at every $X$
(the test operator is pointwise brighter than the family), then
$\mathbb{E}[\hat{X}_\pi(y) - X \mid y] \ge 0$ at every observation level, strictly
where the curves actually separate; the opposite domination flips the sign.
(c) The level-wise bias is
$\delta(y) = \mathbb{E}_{\pi(\theta \mid y)}[g_\theta^{-1}(y)] - g_{\theta^*}^{-1}(y)$.
It vanishes at every level where the posterior mass sits on curves agreeing with
the test curve, and in particular wherever the test curve coincides with a family
member.
\end{proposition}

\begin{proof}[Proof of (a) (tower property)]
$\mathbb{E}[X - \mathbb{E}[X \mid Y]] = 0$ is one line. Its content is empirical
rather than mathematical: a Bayes estimator has no sign bias when the model is
correct, so an in-span bias of $b \approx 0$ is a testable prediction, and the
seen-operator rows of Table~\ref{tab:grid-bias} (Appendix~\ref{app:grid}) are its
test.
\end{proof}

\begin{proof}[Proof of (b), unclipped region ($y < 1$)]
The observation is deterministic, so the posterior over $X$ given $(y, \theta)$ is
the point $g_\theta^{-1}(y)$, and the posterior over $X$ given $y$ is supported on
$\{ g_\theta^{-1}(y) : \theta \in \mathrm{supp}\,\pi \}$. The truth is
$X = g_{\theta^*}^{-1}(y)$. Pointwise domination of the forward maps reverses under
inversion, $g_{\theta^*}^{-1}(y) \le g_\theta^{-1}(y)$ for all $\theta$, hence
\[
\hat{X}_\pi(y) - X
= \mathbb{E}_{\pi(\theta \mid y)}\big[ g_\theta^{-1}(y) \big] - g_{\theta^*}^{-1}(y)
\ge 0,
\]
and the inequality holds for any posterior weights $\pi(\theta \mid y)$: this is
inverse-image dominance, independent of how the model reweights operators after
seeing the data.
\end{proof}

\begin{proof}[Proof of (b), clipped region ($y = 1$)]
The model posterior is the prior truncated from below at
$c_\theta = g_\theta^{-1}(1)$, mixed over $\theta$; the truth satisfies
$X \ge c_{\theta^*}$ with $c_{\theta^*} \le c_\theta$, since a brighter operator
saturates earlier in scene units. The truncated mean
$\mathbb{E}_F[X \mid X \ge c]$ is monotone increasing in the threshold $c$; for a
Gaussian prior it has the closed form
\[
\mathbb{E}[X \mid X \ge c] = \mu + \sigma \cdot \frac{\varphi(\alpha)}{1 - \Phi(\alpha)},
\qquad \alpha = (c - \mu)/\sigma,
\]
monotone by the monotonicity of the hazard function. The model's truncated mean
therefore dominates the true conditional mean. The unclipped and clipped regions
carry the same sign, which explains the frame-consistent direction of the observed
bias.
\end{proof}

\begin{proof}[Proof of (c)]
Immediate from the expression
$\delta(y) = \mathbb{E}_{\pi(\theta \mid y)}[g_\theta^{-1}(y)] - g_{\theta^*}^{-1}(y)$:
at levels $y$ where the test curve coincides with the family, $\delta(y) = 0$. For
\texttt{hard2} the mismatch is concentrated on two axes, the mid-tone slope
($\gamma = 0.6$ versus family slope $1.0$) and the exposure setpoint ($0.35$ versus
$0.18$), so the corollary predicts that residual error localizes in the mid-tone
and shadow levels governed by those axes. This is the localization finding of \S5
applied to ourselves, with no competitor-gap framing. One caution applies:
\texttt{hard2} sets exposure per clip by an auto-exposure target, so in scene units
its curves need not dominate the family pointwise on every clip; part (b)'s
sufficient condition is then not guaranteed there, and the measured \texttt{hard2}
signature is indeed sign-mixed across levels: the level-resolved bias falls from
$+0.92$ in the shadows to $-0.06$ in the highlights (\S5), while the clip-averaged
bias reported in Table~\ref{tab:grid-bias} stays positive. We therefore treat part (b)
as untested on \texttt{hard2} and rely on parts (a) and (c) for the empirical
connection.
\end{proof}

\begin{remark}[General form]
\label{rem:misspec-general}
On \texttt{mustgen} the operator is seen but the content is out of distribution,
and a nonzero directional error remains under the plain L2 stack. Statement (b)
generalizes:
$b = \int (\text{assumed posterior mean} - \text{true posterior mean})\,
dP_{\mathrm{true}}(y)$,
so any misspecified posterior factor produces a directional shift on the luminance
coordinate. Operator out-of-span is the case we solve analytically including the
sign; for content misspecification the sign is not analytically resolved, and we
claim nothing beyond the empirical observation.
\end{remark}

\subsection{Region-dependent optimum of the asymmetry}

\begin{proposition}[Region-dependent optimum of the asymmetry]
\label{prop:gate}
Consider two token types. Observed tokens ($o = 1$, $Y < 1$) have a posterior
concentrated and symmetric around $g_\theta^{-1}(y)$ with small standard deviation
$\sigma_{\mathrm{obs}}$. Censored tokens ($o = 0$) have a lower-truncated posterior
with large $\sigma_{\mathrm{clip}}$ and a possible misspecification shift $\Delta$.
Then: (i) the observed-token risk $R_{\mathrm{obs}}(\tau)$ is uniquely minimized at
$\tau = 1/2$; any $\tau \ne 1/2$ adds the bias term
$\sigma_{\mathrm{obs}} \cdot b(\tau)$ and is a pure loss, where $b(\tau)$ is the
standardized expectile shift, negative for $\tau < 1/2$. (ii) The censored-token
risk $R_{\mathrm{clip}}(\tau)$ is minimized at the $\tau^* < 1/2$ solving
$\sigma_{\mathrm{clip}} \cdot b(\tau^*) = -\Delta$ when $\Delta > 0$: the asymmetry
cancels the bias. (iii) A global constant $\tau$ is a compromise optimal in neither
region, while the gate \texttt{tok\_w} $= 1 + (\lambda - 1)(1 - o)$ attains both
per-region optima ($1/2$ and $\tau^*$) simultaneously at $o \in \{0, 1\}$,
dominating any global constant in risk.
\end{proposition}

\begin{proof}
(i) For a well-specified symmetric posterior,
$e_\tau = \mu + \sigma_{\mathrm{obs}} b(\tau)$ and
$\mathrm{MSE}(\tau) = \sigma_{\mathrm{obs}}^2 + \sigma_{\mathrm{obs}}^2 b(\tau)^2$;
$b(1/2) = 0$ and $b$ is monotone in $\tau$ (property (ii) stated alongside Proposition~\ref{prop:expectile}),
so the minimum is unique. (ii) With shift $\Delta$ the bias is
$\Delta + \sigma_{\mathrm{clip}} b(\tau)$; monotonicity and continuity of $b$ give
existence and uniqueness of the root $\tau^*$. (iii) is the direct combination of
(i) and (ii).
\end{proof}

\begin{remark}[Why observed tokens are immune]
\label{rem:gate-immune}
The unclipped derivation above confines the posterior of an observed token to the
convex hull of $\{ g_\theta^{-1}(y) \}$ regardless of the posterior weights and
regardless of the content prior: the likelihood pins the value. The censored token
has no such pinning; there the prior is everything, so the misspecification shift
$\Delta$ lives mainly on censored tokens, which is the premise of (ii). One honest
exception follows from Proposition~\ref{prop:misspec}(b): when the operator itself
is out of span, observed tokens also carry the bias $\delta(y)$, and this component
cannot be fixed at training time because the training distribution does not contain
that operator. The residual documented in \S5 is exactly this observed-token share
that the asymmetry cannot repair.
\end{remark}

\begin{remark}[Scope]
\label{rem:gate-scope}
Proposition~\ref{prop:gate} is a statement about training risk, where the operator
is in-span by definition. Transfer of the learned asymmetry to test-time mismatch
is a qualitative argument, justified by the distribution-independence of the
expectile direction (Proposition~\ref{prop:expectile}), and is outside the formal
scope; we state this rather than claim it.
\end{remark}

%% file: supp/D_implementation.tex
\section{Implementation Details}
\label{app:impl}

This appendix records the implementation facts referenced from Section~4.

\paragraph{Naming}
The main paper names operators by what they do; the code and the tables below use
the configuration keys. Table~\ref{tab:opnames} maps the two.

\begin{table}[t]
\centering\footnotesize
\begin{tabular}{llc}
\toprule
Main-paper name & Configuration key & In training family \\
\midrule
Hard-Clip & \texttt{gamma\_clip} (\texttt{mustgen}) & yes \\
ACES      & \texttt{aces\_approx}   & yes \\
Reinhard  & \texttt{reinhard\_wp8}  & yes \\
Grading   & \texttt{eval\_profile}  & yes \\
---       & \texttt{train\_profile} & yes \\
Filmic    & \texttt{filmic\_hable}  & no \\
Severe    & \texttt{gamma\_clip\_hard2} & no \\
\bottomrule
\end{tabular}
\caption{Operator names used in the main paper and the corresponding
configuration keys used in the code and in the tables of this supplement.
\texttt{train\_profile} is a training-family member with no evaluation column;
Appendix~\ref{app:grid} explains why.}
\label{tab:opnames}
\end{table}

\paragraph{Operator family}
Five operators synthesize training SDR: \texttt{train\_profile}, \texttt{eval\_profile}, \texttt{gamma\_clip} (pure hard clip; the information-destroying extreme; the mustgen benchmark uses this operator with \texttt{ev\_push} $=0$), \texttt{reinhard\_wp8} \citep{reinhard}, \texttt{aces\_approx} \citep{aces}. The family's auto-exposure setpoint is $0.18$ and its mid-tone slope is $1.0$. Per-operator definitions follow, taken from the synthesis code. Every member except \texttt{train\_profile} first applies per-clip auto-exposure, scaling linear radiance so the positive-luminance median meets the setpoint $0.18$, then a tone curve, then the sRGB OETF, then a clip to $[0,1]$. \texttt{eval\_profile} is Reinhard $x/(1+x)$. \texttt{train\_profile} is luminance-preserving Reinhard with white point $4.0$, at a fixed exposure of $1.0$, and is the one member with no auto-exposure. \texttt{gamma\_clip} uses the identity curve, so the clip is the only nonlinearity. \texttt{reinhard\_wp8} is Reinhard with white point $8.0$. \texttt{aces\_approx} is the Narkowicz ACES approximation. Two operators are held out. \texttt{filmic\_hable} is the Hable curve normalized to white $11.2$ at the same setpoint $0.18$. \texttt{gamma\_clip\_hard2} uses setpoint $0.35$ with a $\gamma = 0.6$ pre-shape applied to linear radiance before the hard clip. The curve-shape and setpoint sampling knobs (\texttt{gamma\_preshape}, \texttt{autoev\_target}) exist in the code but were left at their defaults $1.0$ and $0.18$ for the corpus used here, so the training family is five fixed operators rather than a sampled distribution.

\paragraph{Held-out operators}
\texttt{filmic\_hable} \citep{hable2010} is the soft held-out operator and \texttt{gamma\_clip\_hard2} is the severe one, with auto-exposure target $0.35$ and a $\gamma = 0.6$ mid-tone pre-shape. Neither appears in training.

\paragraph{Observability $o$}
Within a clip, log radiance is binned into $48$ bins, the median SDR level is taken in each bin, and the slope of that empirical curve is read:
\[
\tilde{o} = \max\!\left(\frac{d\,(\text{median SDR})}{d \log_{10} Y},\, 0\right).
\]
The slope is then min-max rescaled across the clip using its $1$st and $99$th percentiles, giving a per-pixel score in $[0,1]$, and mean-pooled inside each latent token to give $o$. Saturated pixels have zero slope and therefore $o = 0$. The estimate assumes no functional form for the operator, and it is read off the training pair, so $o$ exists only during training.

\paragraph{Probe}
The probe is a fixed linear direction $w$ fitted on exposure sweeps of the backbone VAE latent, with held-out $R^2 \approx 0.9$. The vector $w_{\mathrm{eff}}$ is $w$ expressed in the trainer's raw latent basis. The probe is frozen during training; no gradient flows into it.

\paragraph{Asymmetric loss}
The luminance direction is
\[
\texttt{lum\_dir} = -\langle e,\, w_{\mathrm{eff}} \rangle ,
\]
a velocity-error sign flip in which a positive value means the prediction is brighter than ground truth. The weight is $\lambda_{\mathrm{over}} = 2.0$ if $\texttt{lum\_dir} > 0$ and $\lambda_{\mathrm{under}} = 1.0$ otherwise, and the token weight is
\[
\texttt{tok\_w} = 1 + (\lambda - 1)(1 - o).
\]
The loss is decode-free and its training cost is negligible.

\paragraph{Training}
We train IC-LoRA \citep{lora,iclora} at rank 64 on the pretrained LTX-2 backbone \citep{ltx2} for 2500 steps with batch size 4, one training run per stack variant. The training corpus contains 347 unique source clips, of which 187 are real captures from HdM-HDR-2014 \citep{hdm2014} and $156+4$ are synthetic renders, expanded to 1{,}735 training pairs by the five-operator family ($347 \times 5$). There is no explicit real-to-synthetic mixture weight; sampling is uniform over the variant files. We use AdamW with learning rate 2e-4 and a linear schedule; LoRA rank 64, alpha 64, dropout 0, and 10 target modules; bf16 with gradient checkpointing. The values tried during development, selected on the validation split, are the over-weight $\lambda_{\mathrm{over}}$ over $\{1.0, 1.5, 2.0, 3.0\}$ with $\lambda_{\mathrm{under}}$ held at $1.0$ (2.0 selected), and LoRA rank over $\{32, 64\}$ (64 selected). Learning rate, step count and batch size were fixed at the values above and not swept. Every run sets seed 42. The trainer calls accelerate's \texttt{set\_seed} before the training loop, which seeds Python's \texttt{random}, NumPy and torch on CPU and CUDA in each process; the validation sampler draws from a \texttt{torch.Generator} seeded with the same value, and the baseline inference scripts default to the same seed. Deterministic cuDNN kernels are not forced, so a rerun reproduces the training trajectory up to kernel-level nondeterminism rather than bitwise. Per-source clip counts are HdM 187 (real) and Polyhaven 156 plus other 4 (synthetic). There is no explicit mixture weight, since sampling is uniform over variant files.

\paragraph{Inference}
Inference is single-stage with the native 8-step distilled sigma schedule. LumiVid \citep{lumivid} is run in its native 8-step regime, stage one only; its second diffusion stage carries the same HDR adapter and is excluded to equalize cost and resolution, not because it lies outside the HDR reconstruction (Appendix~\ref{app:baselines}).

\paragraph{Scoring}
We score with ColorVideoVDP \citep{cvvdp} using a 1000-nit BT.709 display configuration. Automatic per-clip photometry is the primary protocol, and no response curve is ever re-fitted. We call \texttt{pycvvdp.cvvdp(}\allowbreak\texttt{display\_name=}\allowbreak\texttt{"standard\_hdr\_linear")} with cvvdp 0.5.4; \texttt{photometry-align=auto} resolves per clip (ARRI: none; UPIQ: \texttt{p99\_luminance}; LiU: mixed by scale ratio).

%% file: supp/E_baselines.tex
\section{Baseline Inference Regimes}
\label{app:baselines}

The main text states only that every baseline runs through its own released code path, single-stage. The details follow.

\begin{itemize}
\item \textbf{LumiVid} \citep{lumivid}, released as the LTX-2.3 HDR IC-LoRA \citep{ltx2,lora,iclora}. We run the 8-step schedule of the public release. The paper itself describes an 11-step, two-stage pipeline; the 8-step distilled schedule is a property of the released checkpoints, not a claim of the paper. The release applies a generic latent spatial upsampler and then a second diffusion stage; that second stage carries the same HDR adapter as the first, so it is part of the HDR reconstruction and not a general-purpose refiner. We run stage one only, which equalizes denoising cost and output resolution across methods and places LumiVid below its published two-stage setting.

\item \textbf{X2HDR} \citep{x2hdr}. The method targets RAW input, but the released code ships a non-RAW path: \texttt{input\_is\_raw=False}, with the SDR frame supplied through the spatial-conditioning input (\texttt{spatial\_images}). We use that path as released; no modification of the method was required for SDR input.

\item \textbf{HDRTVNet++} \citep{hdrtvnetpp} and \textbf{LEDiff} \citep{lediff} run at their default published settings.

\item All methods receive the identical SDR input produced by each evaluation operator, at the same resolution and frame count, and are scored by the same photometry protocol.
\end{itemize}

%% file: supp/F_grid.tex
\section{The Full 7-Operator Evaluation Grid}
\label{app:grid}

Tables~\ref{tab:grid-cvvdp} and~\ref{tab:grid-bias} report the full model-by-operator grid on the 84-clip evaluation set (ARRI 42 / UPIQ 30 / LiU HDRv 12), CVVDP with the 1000-nit BT.709 display configuration \citep{cvvdp}, under the automatic per-clip photometry policy. Six operator columns are usable. The seventh training-family operator, \texttt{train\_profile}, is excluded as pathological: it uses a fixed exposure of 1.0 with no auto-exposure, 28 of 84 clips exceed 90\% highlight clipping under it, and the models collapse to CVVDP 2.0--4.4; it is a pathological extreme of the training distribution, not a benchmark. Our final model leads on all operators except \texttt{eval\_profile}, the profile our synthesis pipeline matched to the released decode post-processing of \citet{lumivid}; the training operator of \citet{lumivid} is not disclosed, so we make no claim about what it was trained on. On \texttt{hard2} the leading variant is the multi-op $+$ asymmetric-loss stack (6.885) with the full stack at 6.879.

\begin{table*}[t]
\centering
\footnotesize
\setlength{\tabcolsep}{4pt}
\begin{tabular}{lcccccc}
\toprule
Model & \begin{tabular}[c]{@{}c@{}}mustgen\\(seen,\\hard clip)\end{tabular}
      & \begin{tabular}[c]{@{}c@{}}filmic\\(unseen,\\soft)\end{tabular}
      & \begin{tabular}[c]{@{}c@{}}hard2\\(unseen,\\severe)\end{tabular}
      & \begin{tabular}[c]{@{}c@{}}aces\\(seen)\end{tabular}
      & \begin{tabular}[c]{@{}c@{}}wp8\\(seen)\end{tabular}
      & \begin{tabular}[c]{@{}c@{}}eval\_profile\\(seen)\end{tabular} \\
\midrule
multi-op only        & 7.623 & 8.505 & 6.852 & 8.017 & 8.485 & 8.545 \\
$+$ asymmetric loss  & 7.654 & 8.521 & \textbf{6.885} & 8.032 & 8.500 & 8.555 \\
ReITM (ours)         & \textbf{7.690} & \textbf{8.567} & 6.879 & \textbf{8.103} & \textbf{8.542} & 8.581 \\
LumiVid      & 6.115 & 8.384 & 6.161 & 7.234 & 8.521 & \textbf{8.749} \\
X2HDR        & 5.839 & 7.690 & 5.373 & 6.821 & 7.607 & 7.471 \\
HDRTVNet++   & 6.028 & 7.403 & 5.586 & 5.613 & 7.290 & 6.882 \\
LEDiff       & 5.025 & 3.261 & 4.787 & 4.459 & 3.423 & 3.221 \\
\bottomrule
\end{tabular}
\caption{CVVDP (JOD, higher is better), 84 clips, automatic per-clip photometry. Bold marks the column leader. The staircase from multi-operator only, to $+$ asymmetric loss, to the full stack is monotone on \texttt{mustgen}, \texttt{filmic}, \texttt{aces}, \texttt{wp8}, and \texttt{eval\_profile}.}
\label{tab:grid-cvvdp}
\end{table*}

\begin{table*}[t]
\centering
\footnotesize
\setlength{\tabcolsep}{5pt}
\begin{tabular}{lcccccc}
\toprule
Model & mustgen & filmic & hard2 & aces & wp8 & eval\_profile \\
\midrule
multi-op only        & $+0.032$ & $+0.029$ & $+0.040$ & $+0.025$ & $+0.020$ & $+0.019$ \\
$+$ asymmetric loss  & $+0.027$ & $+0.026$ & $+0.038$ & $\mathbf{+0.013}$ & $\mathbf{+0.014}$ & $\mathbf{+0.012}$ \\
ReITM (ours)         & $+0.026$ & $+0.024$ & $+0.037$ & $+0.018$ & $+0.016$ & $+0.015$ \\
LumiVid      & $+0.089$ & $+0.011$ & $+0.088$ & $-0.013$ & $+0.002$ & $-0.003$ \\
X2HDR                & $+0.079$ & $-0.011$ & $+0.119$ & $-0.042$ & $-0.006$ & $-0.010$ \\
HDRTVNet++           & $-0.047$ & $-0.097$ & $+0.008$ & $-0.261$ & $-0.106$ & $-0.113$ \\
LEDiff               & $+0.016$ & $-0.014$ & $+0.013$ & $-0.058$ & $-0.017$ & $-0.021$ \\
\bottomrule
\end{tabular}
\caption{Signed luminance bias $b$ after the median-luminance alignment of Section~3 (closer to 0 is better; $+$ = too bright), 84 clips, three decimals. This is the full-precision form of the bias values printed to two decimals in main-text Table~1. Against multi-operator training alone, the asymmetric loss reduces $|b|$ on all six operators ($0.032 \to 0.027$, $0.029 \to 0.026$, $0.040 \to 0.038$, $0.025 \to 0.013$, $0.020 \to 0.014$, $0.019 \to 0.012$), which is the claim made in Section~5. Bold marks the smallest $|b|$ among our three arms in each column. Proposition~\ref{prop:misspec}(a) predicts $b=0$ in span; the seen-operator columns of the marginalized rows are small and positive rather than zero, so the prediction holds in sign and magnitude but not exactly, which is consistent with the residual between-operator term reported in Appendix~\ref{app:singleop}.}
\label{tab:grid-bias}
\end{table*}

%% file: supp/G_singleop.tex
\section{Single-Operator Control Arms, and Why the Decomposition is Taken on the Luminance Error}
\label{app:singleop}

\S5 reports a controlled ablation of operator marginalization. Two arms were trained on a single operator each, \texttt{gamma\_clip} (a hard clip) and \texttt{aces\_approx}. They share backbone (LTX-2 \citep{ltx2} + IC-LoRA \citep{lora,iclora} rank 64, frozen backbone), source clips, optimization steps (2{,}500), batch size, learning rate, seed and loss with the multi-operator-only arm; the only difference is the operator distribution, which is also what changes the number of distinct training pairs (347 per single-operator arm against 1{,}735). Step count rather than pair count was matched, because the pair count is a consequence of the treatment. Grid: 84 clips (ARRI 42 / UPIQ 30 / LiU 12) $\times$ 6 operators. Bootstrap: 2{,}000 clip-paired resamples.

\subsection{The decomposition reported in the main text (luminance error)}
\label{app:singleop:lum}

Eq.~(2) is a statement about the reconstruction error, so the decomposition is taken on the luminance error $e$ of \S3: $\mathrm{within} = \mathbb{E}_{T}[\mathrm{Var}(e \mid T)]$, $\mathrm{between} = \mathrm{Var}_{T}(\mathbb{E}[e \mid T])$.

\begin{table*}[t]
\centering
\scriptsize
\caption{Within/between decomposition of the luminance error $e$. Bootstrap intervals are 2{,}000 clip-paired resamples.}
\label{tab:singleop-lum}
\begin{tabular}{lccccc}
\toprule
arm & mean $L$ & spread (max$-$min) & between & between 95\% CI & between / total \\
\midrule
single-op (hard clip) & 0.217 & 0.164 & 0.00326 & [0.00187, 0.00578] & 2.72\% \\
single-op (ACES) & 0.224 & 0.160 & 0.00521 & [0.00401, 0.00670] & 8.03\% \\
\textbf{multi-operator only} & \textbf{0.146} & \textbf{0.048} & \textbf{0.00005} & [0.00003, 0.00010] & \textbf{0.16\%} \\
ReITM (ours) & 0.141 & 0.050 & 0.00006 & [0.00003, 0.00013] & 0.19\% \\
\citet{lumivid} (reference; training operator not disclosed) & 0.179 & 0.138 & 0.00182 & [0.00151, 0.00219] & 4.40\% \\
\bottomrule
\end{tabular}
\end{table*}

Both single-operator arms carry a larger between-operator term than multi-operator
training, with disjoint intervals. The grid mixes four operators inside the training
family with two outside it, so this measures the aggregate reduction in operator
dependence rather than the support restriction of Lemma~\ref{lem:marg}; the
boundary analysis of \S5 isolates the out-of-family axes separately. The two arms are within 0.007 of each other in mean $L$, so we make no claim about which single operator is the better choice.

\subsection{The pre-declared statistic and its degeneracy}
\label{app:singleop:pu21}

The experiment was pre-registered against a different statistic: the same within/between split taken on per-clip PU21 \citep{pu21} error. Under that statistic the declared hypothesis is refuted for one of the two arms, and we report it.

\begin{table}[t]
\centering
\scriptsize
\caption{Within/between decomposition taken on per-clip PU21 error, the pre-declared statistic.}
\label{tab:singleop-pu21}
\setlength{\tabcolsep}{3.5pt}
\begin{tabular}{lccc}
\toprule
arm & within & between & $95\%$ CI \\
\midrule
single-op (hard clip) & 92.87 & 1.70 & [0.71, 5.87] \\
single-op (ACES) & 77.88 & 21.49 & [15.08, 30.44] \\
multi-operator only & 79.11 & 4.74 & [2.29, 9.97] \\
\citet{lumivid} (reference) & 73.21 & 28.25 & [21.00, 37.49] \\
\bottomrule
\end{tabular}
\end{table}

The hard-clip arm has a \emph{smaller} PU21 between-operator term than multi-operator training (1.70 against 4.74), which contradicts the hypothesis. The per-operator means show why.

\begin{table*}[t]
\centering
\scriptsize
\caption{Per-operator mean PU21 (dB) for each arm.}
\label{tab:singleop-permop}
\begin{tabular}{lccccccc}
\toprule
mean PU21 & \texttt{mustgen} & \texttt{aces} & \texttt{wp8} & \texttt{eval\_profile} & \texttt{filmic} & \texttt{hard2} & mean \\
\midrule
single-op (hard clip) & 31.15 & 29.81 & 29.51 & 29.31 & 32.31 & 28.89 & 30.16 \\
single-op (ACES) & 22.42 & 33.85 & 28.58 & 27.90 & 28.31 & 21.12 & 27.03 \\
multi-operator only & 28.88 & 30.00 & 33.05 & 33.39 & 32.23 & 28.41 & 30.99 \\
\bottomrule
\end{tabular}
\end{table*}

The hard-clip arm is uniformly mediocre, spanning 28.9 to 32.3 dB, and its own training operator is only 2.3 dB above its worst cell. A flat and low profile minimizes the variance of per-operator means just as effectively as genuine robustness does, so a between-operator term computed on a quality score cannot separate ``equally good everywhere'' from ``equally mediocre everywhere''. The ACES arm is the opposite case: it is sharply peaked on its own operator, which raises the variance of the means. The luminance error does not have this degeneracy, because a uniformly poor arm simply has a large mean $L$. We therefore report Appendix~\ref{app:singleop:lum} in the main text and record Appendix~\ref{app:singleop:pu21} here rather than choosing the statistic after seeing the result: the decomposition belongs on the error that Eq.~(2) is written for, and the PU21 version is reported because it was the pre-declared one.

\subsection{Why a controlled arm rather than a baseline comparison}
\label{app:singleop:replaces}

Comparing the between-operator term of \citet{lumivid} against ours would confound the operator distribution with architecture, training data and loss, and it would require assuming that \citet{lumivid} was trained on a single operator, which its paper does not state. The arms above remove both problems. \citet{lumivid} is retained in Appendix~\ref{app:singleop:lum} only as a reference point, with no attribution of its value to an operator count.

\subsection{Per-operator cells of the control arms}
\label{app:singleop:cells}

Table~\ref{tab:singleop-cells} opens the aggregates of
Appendix~\ref{app:singleop:lum} into their six operator cells. Each arm attains its
lowest luminance error on the operator it was trained on, and its bias there is the
closest to zero it achieves anywhere. On its worst other column the error grows by a
factor of $2.0$ to $2.1$, and averaged over the five other columns by $1.3$ to $1.7$.
The bias changes sign between the two arms: the hard-clip arm
runs dark on ACES ($-0.155$) while the ACES arm runs bright on the hard clip
($+0.147$). The direction of the residual is therefore set by the training operator,
which is the effect Eq.~(2) of the main paper attributes to the between-operator
term. A single-operator arm buys a better bias on its own operator than either
marginalized stack achieves there ($-0.005$ against $+0.032$ for multi-operator
training and $+0.026$ for the full stack); it pays for that cell with the degradation
in every other column.

\begin{table*}[t]
\centering\footnotesize
\caption{Per-operator luminance error $L$ (RMSE, lower better) and signed bias $b$
($+$ = too bright) for the two single-operator control arms, the multi-operator-only
arm and the full stack, on the six-operator grid over $84$ clips under the
median-luminance alignment of Section~3. The training operator of each control arm is
set in bold. Filmic and Severe are outside the five-operator family and so are unseen
for every row; for the two control arms the other four columns are unseen as well. The
control-arm rows and the two marginalized rows come from two scoring runs on the same
grid, clip set and alignment.}
\label{tab:singleop-cells}
\begin{tabular}{lcccccc}
\toprule
 & \multicolumn{2}{c}{Hard-Clip (\texttt{gamma\_clip})} & \multicolumn{2}{c}{ACES} & \multicolumn{2}{c}{Reinhard (\texttt{wp8})} \\
\cmidrule(lr){2-3}\cmidrule(lr){4-5}\cmidrule(lr){6-7}
arm & $L$ & $b$ & $L$ & $b$ & $L$ & $b$ \\
\midrule
single-op (hard clip) & \textbf{0.173} & \textbf{$-$0.005} & 0.337 & $-$0.155 & 0.190 & $-$0.055 \\
single-op (ACES)      & 0.269 & $+$0.147 & \textbf{0.141} & \textbf{$+$0.001} & 0.205 & $+$0.028 \\
multi-operator only   & 0.154 & $+$0.032 & 0.160 & $+$0.025 & 0.126 & $+$0.020 \\
ReITM (ours)          & 0.149 & $+$0.026 & 0.153 & $+$0.018 & 0.123 & $+$0.016 \\
\midrule
 & \multicolumn{2}{c}{Grading (\texttt{eval\_profile})} & \multicolumn{2}{c}{Filmic (unseen)} & \multicolumn{2}{c}{Severe (unseen)} \\
\cmidrule(lr){2-3}\cmidrule(lr){4-5}\cmidrule(lr){6-7}
arm & $L$ & $b$ & $L$ & $b$ & $L$ & $b$ \\
\midrule
single-op (hard clip) & 0.193 & $-$0.057 & 0.231 & $-$0.068 & 0.178 & $+$0.028 \\
single-op (ACES)      & 0.212 & $+$0.018 & 0.219 & $+$0.021 & 0.301 & $+$0.188 \\
multi-operator only   & 0.125 & $+$0.019 & 0.135 & $+$0.029 & 0.173 & $+$0.040 \\
ReITM (ours)          & 0.121 & $+$0.015 & 0.132 & $+$0.024 & 0.170 & $+$0.037 \\
\bottomrule
\end{tabular}
\end{table*}

%% file: supp/H_moved.tex
\section{Expanded Forms of Main-Text Tables}
\label{app:moved}

\paragraph{Per-source breakdown of CVVDP}
The main text prints this breakdown in compressed slash-separated form. It is
repeated here one cell per column so the nine source-by-benchmark values can be
read directly. CVVDP JOD by data source; ARRI $n=42$, UPIQ $n=30$, LiU $n=12$ per benchmark.
ReITM attains the best score in all nine source-by-benchmark cells, so the
aggregate results are not driven by any single source. UPIQ is the hardest
source for every method except X2HDR, whose weakest source is LiU on all
three benchmarks.

\begin{table*}[t]
\centering
\footnotesize
\setlength{\tabcolsep}{4pt}
\begin{tabular}{lccccccccc}
\toprule
 & \multicolumn{3}{c}{Hard-Clip} & \multicolumn{3}{c}{Filmic} & \multicolumn{3}{c}{Severe} \\
\cmidrule(lr){2-4}\cmidrule(lr){5-7}\cmidrule(lr){8-10}
Method (CVVDP JOD $\uparrow$) & ARRI & UPIQ & LiU & ARRI & UPIQ & LiU & ARRI & UPIQ & LiU \\
\midrule
\textbf{ReITM (ours)} & 8.63 & 6.35 & 7.76 & 9.28 & 7.57 & 8.57 & 7.98 & 5.06 & 7.54 \\
LumiVid    & 6.72 & 5.01 & 6.78 & 9.21 & 7.31 & 8.17 & 6.94 & 4.84 & 6.75 \\
HDRTVNet++ & 7.30 & 4.64 & 5.06 & 8.19 & 6.24 & 7.55 & 7.52 & 2.74 & 5.93 \\
X2HDR      & 7.06 & 5.14 & 3.34 & 8.98 & 7.00 & 4.90 & 6.78 & 4.50 & 2.62 \\
LEDiff     & 7.05 & 2.01 & 5.48 & 5.47 & 0.21 & 3.15 & 6.97 & 1.58 & 5.16 \\
\bottomrule
\end{tabular}
\caption{Per-source breakdown of CVVDP JOD.}
\label{tab:moved-cvvdp-source}
\end{table*}

\paragraph{Fidelity: PU21-PSNR (dB, source-averaged)}
The main text prints these values paired with CVVDP in a single cell. They are
separated here. Same clips and same alignment as the CVVDP table, but averaged
per source and then over the three sources, as in the main text; ReITM leads every
benchmark under a second, magnitude-based metric.

\begin{table}[htbp]
\centering
\footnotesize
\begin{tabular}{lccc}
\toprule
Method (PU21-PSNR, dB $\uparrow$) & Hard-Clip & Filmic & Severe \\
\midrule
\textbf{ReITM (ours)} & 28.07 & 31.69 & 27.66 \\
LumiVid    & 21.67 & 31.20 & 23.45 \\
HDRTVNet++ & 24.95 & 29.06 & 26.82 \\
X2HDR      & 23.27 & 30.68 & 22.36 \\
LEDiff     & 23.36 & 20.99 & 23.23 \\
\bottomrule
\end{tabular}
\caption{Fidelity measured by PU21-PSNR (dB), averaged over sources.}
\label{tab:moved-pu21psnr}
\end{table}

\paragraph{Boundary self-diagnosis on Severe}
Figure~6 of the main paper reports this diagnosis and is not repeated here. The
level-resolved luminance error falls from $+0.92$ in the shadows to $-0.06$ in the
highlights, so the profile is sign-mixed across levels.
Proposition~\ref{prop:misspec}(c) predicts only that the residual localizes on the
levels governed by the two out-of-family axes; the sign is fixed by part~(b) under
pointwise domination, which we do not verify for this operator. The
multi-operator-only stack shows the same profile.

\subsection{Per-cell significance of the luminance error}
\label{app:moved:sig}

Table~1 of the main paper marks significance with an asterisk per cell and reports
that our lead holds in $22$ of $24$ method-by-operator comparisons and is significant
in $18$. Table~\ref{tab:cellsig} gives the underlying numbers. Each entry is
$\Delta = L_{\text{baseline}} - L_{\text{ReITM}}$, so positive favours us, with a
clip-paired bootstrap $95\%$ interval over the $84$ clips and a Wilcoxon signed-rank
$p$-value. We call a cell significant when the interval excludes zero.

Six cells are not significant. Two of them, Reinhard and Grading against
\citet{lumivid}, have a negative point estimate, so the baseline is nominally ahead;
both intervals contain zero. The remaining four are cells where we lead but the
interval reaches zero. No baseline lead in the grid has an interval excluding zero.

\begin{table*}[t]
\centering\scriptsize
\setlength{\tabcolsep}{3pt}
\caption{$\Delta = L_{\text{baseline}} - L_{\text{ReITM}}$ on log-luminance RMSE,
positive favouring us, with clip-paired bootstrap $95\%$ intervals ($84$ clips) and
Wilcoxon signed-rank $p$. Cells whose interval excludes zero are marked $^{*}$.}
\label{tab:cellsig}
\begin{tabular}{llccc@{\hskip 14pt}llccc}
\toprule
Operator & Baseline & $\Delta$ & $95\%$ CI & $p$ & Operator & Baseline & $\Delta$ & $95\%$ CI & $p$ \\
\midrule
\multirow{4}{*}{Hard-Clip} & LumiVid & $+0.1026^{*}$ & $[+0.0819, +0.1228]$ & $7.2\times10^{-13}$ & \multirow{4}{*}{Grading} & LumiVid & $-0.0070$ & $[-0.0213, +0.0046]$ & $1.8\times10^{-1}$ \\
 & X2HDR & $+0.0650^{*}$ & $[+0.0450, +0.0857]$ & $2.6\times10^{-9}$ &  & X2HDR & $+0.0212^{*}$ & $[+0.0049, +0.0357]$ & $3.6\times10^{-6}$ \\
 & HDRTVNet++ & $+0.1042^{*}$ & $[+0.0842, +0.1240]$ & $4.9\times10^{-13}$ &  & HDRTVNet++ & $+0.0912^{*}$ & $[+0.0699, +0.1122]$ & $1.3\times10^{-11}$ \\
 & LEDiff & $+0.1614^{*}$ & $[+0.1429, +0.1800]$ & $2.6\times10^{-15}$ &  & LEDiff & $+0.2552^{*}$ & $[+0.2354, +0.2760]$ & $1.7\times10^{-15}$ \\
\cmidrule(lr){1-5}\cmidrule(lr){6-10}
\multirow{4}{*}{ACES} & LumiVid & $+0.0815^{*}$ & $[+0.0606, +0.1014]$ & $2.1\times10^{-9}$ & \multirow{4}{*}{Filmic (unseen)} & LumiVid & $+0.0162$ & $[-0.0031, +0.0317]$ & $2.4\times10^{-7}$ \\
 & X2HDR & $+0.0526^{*}$ & $[+0.0348, +0.0691]$ & $4.0\times10^{-7}$ &  & X2HDR & $+0.0148$ & $[-0.0082, +0.0338]$ & $3.4\times10^{-5}$ \\
 & HDRTVNet++ & $+0.3230^{*}$ & $[+0.2735, +0.3749]$ & $2.3\times10^{-15}$ &  & HDRTVNet++ & $+0.1456^{*}$ & $[+0.1197, +0.1726]$ & $1.4\times10^{-14}$ \\
 & LEDiff & $+0.1512^{*}$ & $[+0.1272, +0.1763]$ & $9.1\times10^{-15}$ &  & LEDiff & $+0.2464^{*}$ & $[+0.2246, +0.2674]$ & $2.5\times10^{-15}$ \\
\cmidrule(lr){1-5}\cmidrule(lr){6-10}
\multirow{4}{*}{Reinhard} & LumiVid & $-0.0048$ & $[-0.0209, +0.0079]$ & $4.8\times10^{-2}$ & \multirow{4}{*}{Severe (unseen)} & LumiVid & $+0.0344^{*}$ & $[+0.0173, +0.0507]$ & $4.3\times10^{-5}$ \\
 & X2HDR & $+0.0140$ & $[-0.0039, +0.0297]$ & $6.6\times10^{-5}$ &  & X2HDR & $+0.0760^{*}$ & $[+0.0557, +0.0977]$ & $1.8\times10^{-9}$ \\
 & HDRTVNet++ & $+0.0773^{*}$ & $[+0.0576, +0.0973]$ & $1.8\times10^{-10}$ &  & HDRTVNet++ & $+0.0033$ & $[-0.0194, +0.0240]$ & $5.9\times10^{-2}$ \\
 & LEDiff & $+0.2446^{*}$ & $[+0.2251, +0.2642]$ & $1.8\times10^{-15}$ &  & LEDiff & $+0.1411^{*}$ & $[+0.1235, +0.1598]$ & $2.5\times10^{-15}$ \\
\bottomrule
\end{tabular}
\end{table*}

\subsection{Region-stratified fidelity}
\label{app:moved:region}

The main text states that restricting PU21-PSNR to SDR luma $\ge 0.96$ raises the
margin over \citet{lumivid}, without giving the numbers. Table~\ref{tab:region}
gives them. Scoring and averaging follow main-text Table~3 exactly: freeze protocol,
$84$ clips, median-luminance alignment, per-source means then averaged over the three
sources. The margin grows on all three benchmarks when the score is restricted to the
near-saturated region, which is the region the forward operator destroyed.

Clip counts differ between the two rows of each benchmark because a clip contributes
to the near-saturated score only if it has such a region. On Filmic that condition
holds for $53$ of $84$ clips, and in $16$ of those the region is a few pixels wide, so
the Filmic restricted column rests on less material than its clip count suggests. The
Filmic gain is also concentrated in one source: UPIQ moves by $+3.86$\,dB while ARRI
and LiU move by under $0.7$\,dB.

The same files carry the signed bias split by region, and there the picture is not
uniformly in our favour. Restricting to SDR luma $\le 0.04$ leaves only $12$ of the
$84$ clips with a shadow region at all, so the following is a small sample. On those
clips, under the same source-averaged convention, our shadow bias on Severe is
$+1.07$ against $+0.10$ for HDRTVNet++. The two other generative methods sit with us
($+1.11$ for LumiVid, $+1.19$ for X2HDR), so this separates regression from
generation rather than us from the field. It is the shadow weakness the boundary
analysis of Section~5 identifies on our own stack, measured here against a
competitor.

\begin{table}[htbp]
\centering\footnotesize
\caption{PU21-PSNR (dB, source-averaged) over the whole frame and restricted to SDR
luma $\ge 0.96$. Protocol identical to main-text Table~3. $n$ counts the clips with a
near-saturated region under our reconstruction; for LumiVid the count is the same on
Hard-Clip and Severe and one higher on Filmic.}
\label{tab:region}
\setlength{\tabcolsep}{3pt}
\begin{tabular}{llccc}
\toprule
Benchmark & Region & \shortstack{ReITM\\(ours)} & LumiVid & margin \\
\midrule
\multirow{2}{*}{Hard-Clip} & full frame & $28.07$ & $21.67$ & $+6.40$ \\
 & luma $\ge 0.96$ ($n=81$) & $19.34$ & $12.00$ & $\mathbf{+7.34}$ \\
\midrule
\multirow{2}{*}{Filmic} & full frame & $31.69$ & $31.20$ & $+0.48$ \\
 & luma $\ge 0.96$ ($n=53$) & $22.13$ & $20.42$ & $\mathbf{+1.72}$ \\
\midrule
\multirow{2}{*}{Severe} & full frame & $27.66$ & $23.45$ & $+4.20$ \\
 & luma $\ge 0.96$ ($n=83$) & $21.94$ & $16.99$ & $\mathbf{+4.95}$ \\
\bottomrule
\end{tabular}
\end{table}

\subsection{The alignment sweep for all five methods}
\label{app:moved:align}

Main-text Table~4 sweeps three photometric-alignment levels for three of the five
methods. Table~\ref{tab:alignfull} completes it. Nothing else changes between the
three levels: the same predictions, the same $84$-clip manifests, the same scoring
call and the same display configuration; only the alignment argument differs.

ReITM is first on every benchmark at every level, nine of nine cases, which is the
claim the main text makes on the partial table. Alignment helps \citet{lumivid} more
than it helps us on all three benchmarks, because a global scale forgives a larger
tone and exposure error; the other three baselines gain less than we do on at least
one benchmark each, so this is a statement about the strongest baseline and not about
baselines in general. Our margin over \citet{lumivid} therefore narrows as alignment
increases, and on Filmic under full median alignment it falls to $+0.096$ JOD. The main text already describes the
Filmic margin as narrow and reports it as not significant under the primary metric;
this is the same fact measured under a different alignment.

Two rows are not monotone in the alignment level. LEDiff is non-monotone on all
three benchmarks and HDRTVNet++ on Severe. LEDiff's outputs collapse in tone and
colour, so a global scale factor cannot recover them, and its three levels differ by
at most $0.32$ JOD, against a gap to us of at least $1.53$ JOD.

\begin{table*}[t]
\centering\footnotesize
\setlength{\tabcolsep}{6pt}
\caption{CVVDP (JOD $\uparrow$) at three photometric-alignment levels: none, the
automatic per-clip policy used as the primary protocol, and full median-luminance.
$84$ clips.}
\label{tab:alignfull}
\begin{tabular}{lccc ccc ccc}
\toprule
 & \multicolumn{3}{c}{Hard-Clip} & \multicolumn{3}{c}{Filmic} & \multicolumn{3}{c}{Severe} \\
\cmidrule(lr){2-4}\cmidrule(lr){5-7}\cmidrule(lr){8-10}
Method & none & auto & med. & none & auto & med. & none & auto & med. \\
\midrule
\textbf{ReITM (ours)} & \textbf{6.98} & \textbf{7.69} & \textbf{7.81} & \textbf{7.61} & \textbf{8.57} & \textbf{8.68} & \textbf{6.43} & \textbf{6.88} & \textbf{6.99} \\
LumiVid    & 5.27 & 6.12 & 6.53 & 7.34 & 8.38 & 8.58 & 5.71 & 6.16 & 6.50 \\
HDRTVNet++ & 5.93 & 6.03 & 6.68 & 7.18 & 7.40 & 7.85 & 5.82 & 5.59 & 5.98 \\
X2HDR      & 5.13 & 5.84 & 6.17 & 7.43 & 7.69 & 7.79 & 4.34 & 5.37 & 5.55 \\
LEDiff     & 5.15 & 5.03 & 5.14 & 3.58 & 3.26 & 3.48 & 4.90 & 4.79 & 4.94 \\
\bottomrule
\end{tabular}
\end{table*}

\subsection{Structural similarity}
\label{app:moved:ssim}

The main paper reports a perceptual metric and a magnitude metric. SSIM and its
PU21-encoded form are a third family, sensitive to structure rather than to level.
Table~\ref{tab:ssim} reports both under median-luminance alignment on the same $84$
clips.

We lead on Hard-Clip under both forms and on Filmic under PU21-SSIM. On Severe, HDRTVNet++ leads
PU21-SSIM, $0.937$ against our $0.911$, while we lead linear SSIM there, $0.692$
against its $0.540$. HDRTVNet++ also leads linear SSIM on Filmic, $0.884$ against our
$0.850$. A regression model that under-expands produces a smooth, well-correlated
image, which structural measures reward and which the luminance error of Section~3
does not: the same model carries a bias of $-0.097$ on Filmic and $-0.261$ on ACES
(Table~\ref{tab:grid-bias}), the largest magnitudes in that table.

\begin{table}[htbp]
\centering\footnotesize
\caption{PU21-SSIM and linear SSIM, median-luminance alignment, $84$ clips,
clip-averaged.}
\label{tab:ssim}
\begin{tabular}{lcc cc cc}
\toprule
 & \multicolumn{2}{c}{Hard-Clip} & \multicolumn{2}{c}{Filmic} & \multicolumn{2}{c}{Severe} \\
\cmidrule(lr){2-3}\cmidrule(lr){4-5}\cmidrule(lr){6-7}
Method & PU21 & lin. & PU21 & lin. & PU21 & lin. \\
\midrule
\textbf{ReITM (ours)} & \textbf{.930} & \textbf{.736} & \textbf{.947} & .850 & .911 & \textbf{.692} \\
LumiVid    & .899 & .345 & .937 & .810 & .899 & .535 \\
HDRTVNet++ & .925 & .617 & .937 & \textbf{.884} & \textbf{.937} & .540 \\
X2HDR      & .906 & .487 & .933 & .796 & .883 & .493 \\
LEDiff     & .832 & .429 & .796 & .238 & .831 & .410 \\
\bottomrule
\end{tabular}
\end{table}

%% file: supp/I_qualitative.tex
\section{Extended Qualitative Comparison}
\label{app:qual}

\begin{figure*}[p]
\centering
\includegraphics[width=\linewidth]{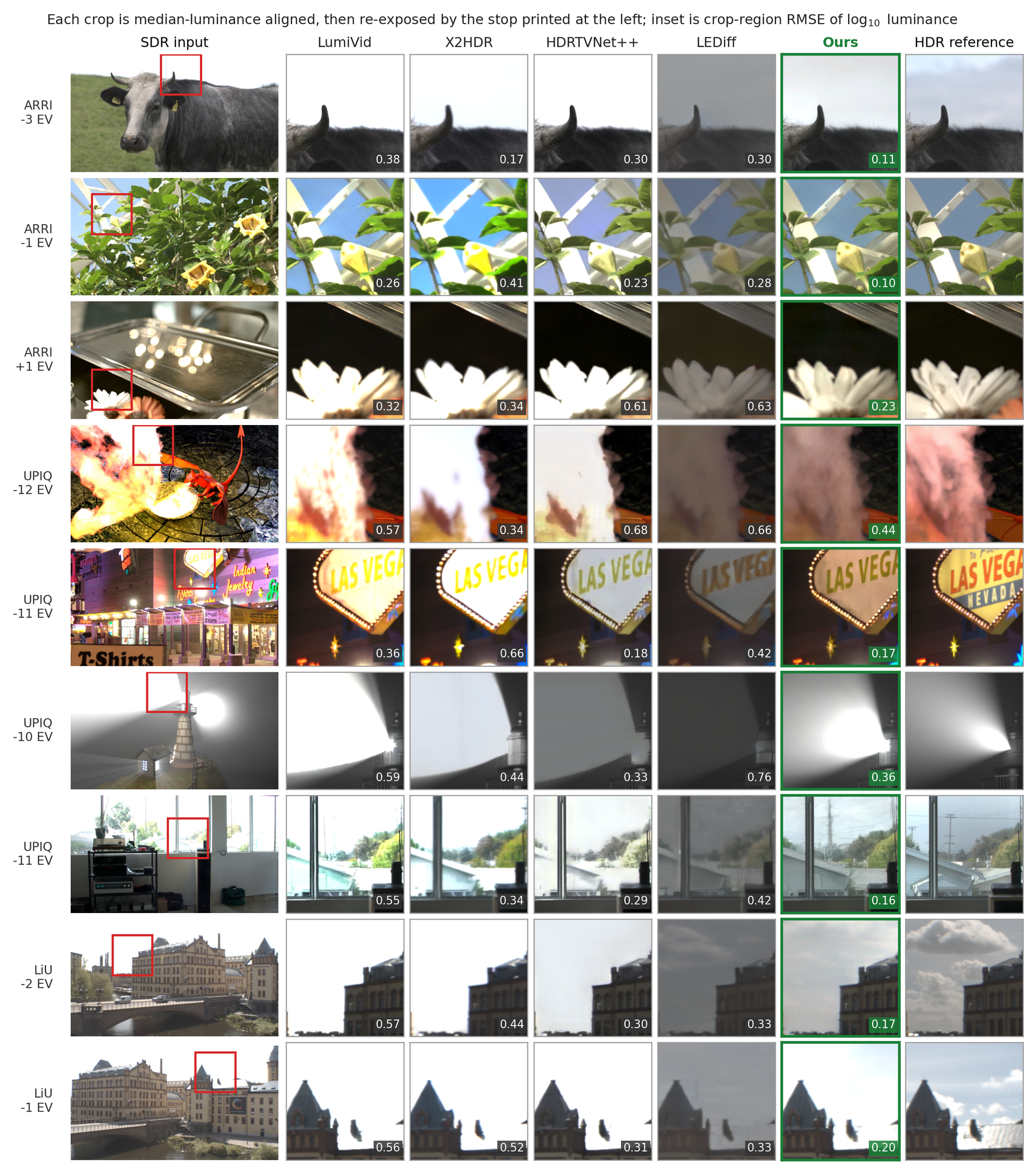}
\caption{Re-exposure of clipped regions under the Hard-Clip operator. Each row
shows the SDR input with the crop marked in red, then that crop for LumiVid,
X2HDR, HDRTVNet++, LEDiff, ReITM (ours) and the reference, all median-luminance aligned,
re-exposed by the stop printed at the left of the row and displayed with $2.2$
gamma. Inset: crop-region RMSE of $\log_{10}$ luminance against the reference.
Rows 1--3 are ARRI, rows 4--7 UPIQ and rows 8--9 LiU HDRv.}
\label{fig:qual-reexp}
\end{figure*}

Figure~\ref{fig:qual-reexp} re-exposes clipped regions. Each row takes one crop
of the SDR input, marked in red, and shows that crop for every method after the
median-luminance alignment of \S5, re-exposed by the single stop printed at the
left of the row. The stop brings the reference crop near the top of the display
range, so a highlight that retains detail separates from one filled with a
constant. The inset is the crop-region RMSE of $\log_{10}$ luminance against the
reference. It reports the luminance error over the crop and carries no judgement
of the content written into a saturated region; the first item of
Appendix~\ref{app:limits} states why no such judgement follows from these scores.

The remainder of this appendix extends the main-text qualitative figure to all
four baselines and to multiple clips per benchmark. Clip selection there is
rule-based, not curated:
clips are taken at fixed percentiles (20/45/65/85) of ReITM's CVVDP
distribution, with at least one non-ARRI source enforced per benchmark. Each
panel shows the signed-luminance error map (blue = too dark, red = too bright;
per-clip median-aligned, $\pm 0.4$ $\log_{10}$); SDR input and ground truth are
shown tone-mapped for reference. Across clips and benchmarks, the baselines'
maps carry coherent color casts in saturated regions, method-specific in sign,
while ReITM remains the most neutral. This is the per-clip visual counterpart of
the luminance-error tables.

\begin{figure*}[!t]
\centering
\includegraphics[width=\linewidth]{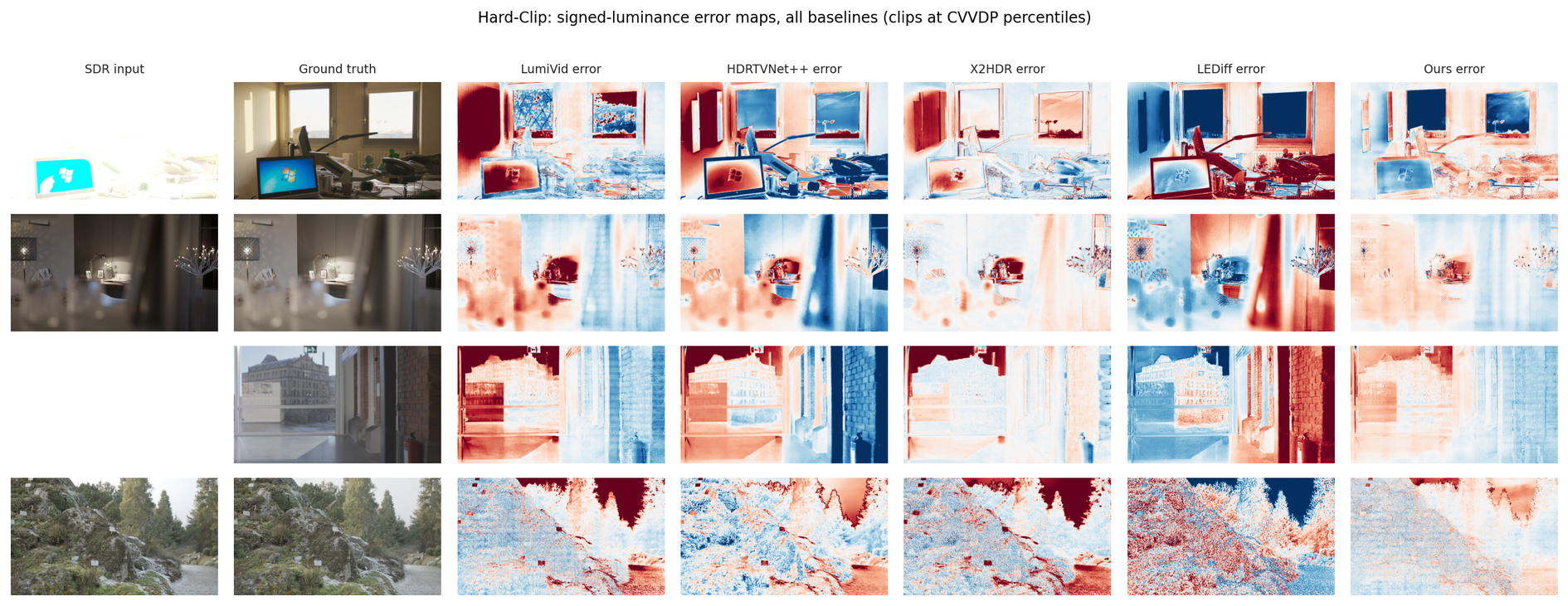}\\[6pt]
\includegraphics[width=\linewidth]{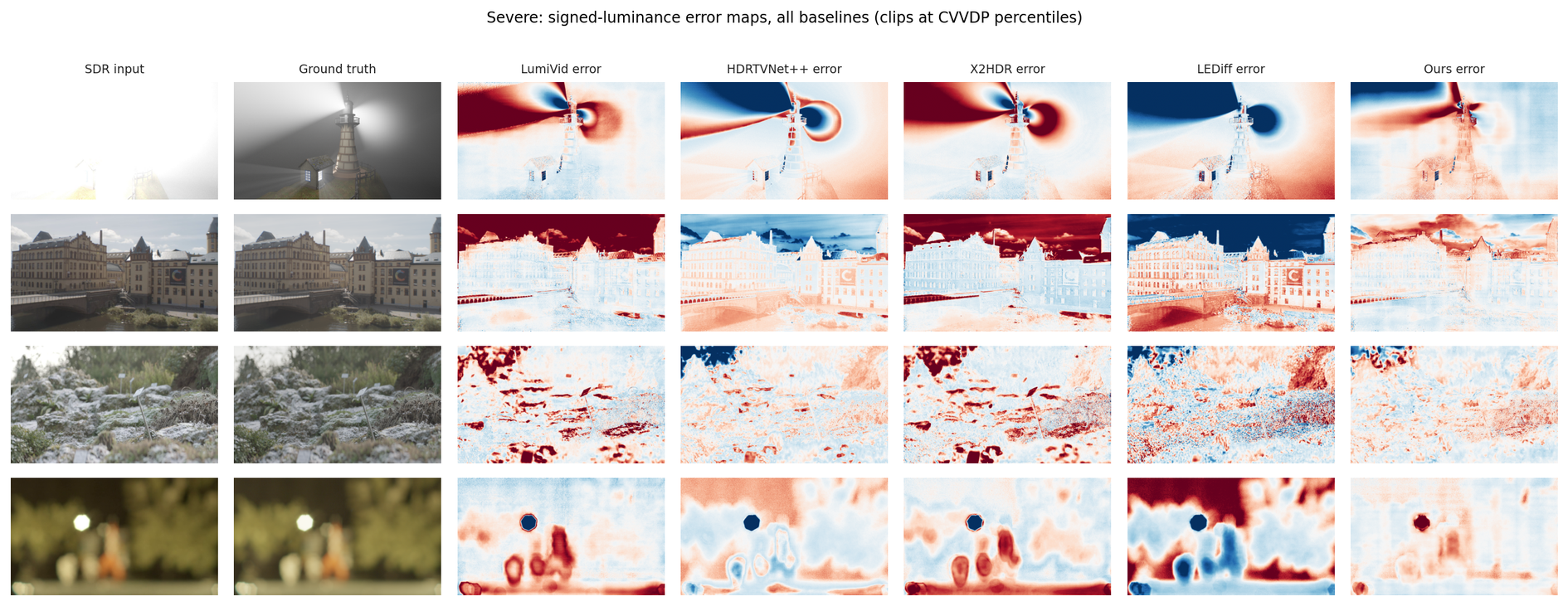}
\caption{Extended qualitative comparison. Each block shows, per clip, the SDR
input, the ground truth, and the signed-luminance error maps for LumiVid,
HDRTVNet++, X2HDR, LEDiff and ReITM (ours). \textbf{Top:} Hard-Clip, clips
\texttt{upiq\_narwaria\_09}, \texttt{arri\_B001C002}, \texttt{liu\_hallway2},
\texttt{arri\_M001C003}. \textbf{Bottom:} Severe, same layout, clips
\texttt{upiq\_narwaria\_07}, \texttt{liu\_bridge\_2}, \texttt{arri\_M001C006},
\texttt{liu\_astronauts}.}
\label{fig:qual-supp}
\end{figure*}

%% file: supp/L_limitations.tex
\section{Limitations}
\label{app:limits}

The main text states these in one sentence; here they are in full. An attempt to
infer the operator at deployment time, which did not work, closes the appendix.

\begin{itemize}
\item \textbf{Generative content is unverifiable.} The particular structure written into a clipped region is one sample from a wide posterior and has no unique correct value; we constrain the direction of the reconstruction, not its content. Region-restricted scores (Appendix~\ref{app:moved}) measure the aggregate error there, not the correctness of any individual detail.

\item \textbf{Marginalization has a support.} The cancellation of the between-operator term holds only over the span of the training family; outside it no cancellation is guaranteed, and the residual measured on Severe is what survives (the range condition tested in \S5). Extending the family is future work.

\item \textbf{Probe scope.} The luminance probe is demonstrated on one backbone with partial cross-backbone transfer ($R^2 = 0.592$), and the held-out validation of Fig.~3 uses computer-generated clips only, because the real-scene clips available to us were used to fit the probe. Table~\ref{tab:probe-sweep} reports the exposure sweep on real clips, one of which shares a scene with the fit set.

\item \textbf{Variance comparison.} The within/between-operator split is now a controlled ablation (Appendix~\ref{app:singleop}), but it rests on two choices of the single training operator, a hard clip and an ACES approximation. We do not claim the effect size is the same for every possible choice.

\item \textbf{Relative output scale.} ReITM, like every baseline, emits radiance on a relative scale and targets no absolute-luminance convention, so every comparison here---and any practical use---requires a photometric scale to be set downstream; the reference sets in our evaluation themselves mix absolute-nit and relative conventions.

\item \textbf{Alignment can move error between regions.} The median-luminance alignment that removes that degree of freedom is a single global factor fitted on the whole frame, so a factor that centers the mid-tones may leave highlights or shadows further from the reference than the unaligned output was. Regional comparisons (Fig.~2 of the main text) are therefore read as examples, not as regional scores.

\item \textbf{Single training run per stack.} Confidence intervals come from per-clip resampling; variance across training seeds is not characterized.
\end{itemize}

\subsection{Operator inference at deployment time}
\label{app:negative}

A natural extension of our framework is to estimate the tone-mapping operator at deployment time from the unclipped pixels of the input and to recalibrate the output accordingly. We built a regressor from unclipped-pixel statistics (luminance quantiles, clip ratio, upper-slope features) to a $K$-knot tone curve, and evaluated several variants: a scalar gamma regressor, a confidence-gated version, per-knot shrinkage toward the identity, and a heteroscedastic v2.

All variants failed to improve over the uncorrected output. On the hard-clip and filmic benchmarks the raw (uncorrected) output scored 7.612 and 8.507 CVVDP respectively, while every non-oracle corrected variant scored between 7.16 and 7.94 (pre-freeze 74-clip internal comparison). An oracle correction using the true operator gained only $+0.1$ to $+0.16$, so the headroom for any output recalibration of this form is small even when the operator is known. One variant with a fixed $\gamma = 2.0$ correction did win, but the true test-operator value was 2.05; the win reflects a coincidental match to the ground truth rather than blind inference, and we do not count it as a contribution. We record this because the approach is the obvious next step a reader would propose. At our current inference accuracy, leaving the output uncorrected is the better decision. No result in the main text depends on this.

%% file: supp/M_reproducibility.tex
\section{Reproducibility Summary}
\label{app:repro}

\paragraph{Model and backbone}
LTX-2 pretrained video diffusion backbone \citep{ltx2}; adaptation via IC-LoRA \citep{lora,iclora} at rank 64. Inference uses the native 8-step distilled sigmas, single stage. The compared \citep{lumivid} checkpoint is run in its native 8-step regime, stage one only, so that both systems are compared at matched denoising cost and output resolution (Appendix~\ref{app:baselines}).

\paragraph{Training}
2500 steps, batch size 4. AdamW, learning rate $2\times10^{-4}$, linear schedule; LoRA rank 64/alpha 64, dropout 0, 10 target modules; bf16 with gradient checkpointing; per-source clips HdM 187 / synthetic 160; no explicit mixture weight (uniform sampling). The backbone's reference recipe used batch size 1 at the same learning rate; this affects absolute comparability to that recipe, not our internal A/B comparisons. One training run per stack variant was performed; we do not report seed variance across training runs, and confidence intervals are per-clip bootstrap at evaluation time.

\paragraph{Operator family (multi-operator synthesis pipeline)}
SDR is synthesized from HDR through five explicit operators: \texttt{train\_profile}, \texttt{eval\_profile}, \texttt{gamma\_clip} (hard clip, information-destroying extreme), \texttt{reinhard\_wp8} \citep{reinhard}, \texttt{aces\_approx} \citep{aces}. Training content combines real HDR (HdM-HDR-2014 \citep{hdm2014}) and synthetic HDR. Unseen test operators, never used in training: \texttt{filmic\_hable} \citep{hable2010} and \texttt{gamma\_clip\_hard2} (auto-exposure target 0.35 versus family 0.18; $\gamma = 0.6$ mid-tone pre-shape versus family slope 1.0).

\paragraph{Evaluation}
84 clips: ARRI 42, UPIQ 30 \citep{upiq}, LiU HDRv 12. Metrics: ColorVideoVDP (JOD) \citep{cvvdp} with a 1000-nit BT.709 display configuration matching the PQ HDR10 grading of the outputs; PU21-PSNR and PU21-SSIM \citep{pu21}; the signed luminance bias $b$, reported whole-frame and restricted to the near-saturated region. Automatic per-clip photometry is the primary protocol; a three-level alignment sweep (none / auto / median) is reported in the main text.

\paragraph{Statistical tests}
Paired method comparisons use 10{,}000-sample clip-paired bootstrap CIs and Wilcoxon signed-rank tests. The within/between variance decomposition of Appendix~\ref{app:singleop} uses 2{,}000 clip-paired resamples, as the main text states.

\paragraph{Compute and environment}
$1\times$ NVIDIA H100 NVL (95{,}830 MiB, driver 550.144.03); 510.5 minutes for the 2{,}500-step training run; peak memory 71.7 GB. Ubuntu 22.04.5 LTS, Linux 6.8.0; Python 3.11.11; PyTorch 2.6.0+cu126 with CUDA 12.6 and cuDNN 9.5.1; bf16 autocast with gradient checkpointing. Scoring uses ColorVideoVDP 0.5.4.

\paragraph{Code and data}
Dataset licenses (HdM-HDR-2014, UPIQ, LiU HDRv) are respected; we redistribute no source HDR content.